\documentclass[11pt,a4paper]{article}
\usepackage[margin=25mm]{geometry}
\usepackage[T1]{fontenc}
\usepackage[utf8]{inputenc}
\usepackage{lmodern,microtype,amsmath,amssymb,booktabs,array,graphicx,xcolor}
\usepackage{longtable,enumitem,caption,placeins,flafter,float,needspace,pdflscape}
\usepackage[numbers,sort&compress]{natbib} \usepackage{xurl}
\usepackage[colorlinks=true,linkcolor=teal,citecolor=teal,urlcolor=teal]{hyperref}
\hypersetup{pdftitle={Various visual cue detection in RGB photographs with forced focused attention}}
\newcommand{\FFA}{\textsc{FFA}}
\newcommand{\R}{\mathbb{R}}

\title{\texttt{vividRGB}: Various visual cue detection in RGB photographs with forced focused attention}
\author{Masahiro Ryo\\[3pt]
\small Leibniz Centre for Agricultural Landscape Research (ZALF)\\
\small Eberswalder Stra\ss e 84, 15374 M\"uncheberg, Germany}
\date{last revised on 24 October 2026}
\begin{document}
\maketitle
\begin{center}
\includegraphics[width=\linewidth]{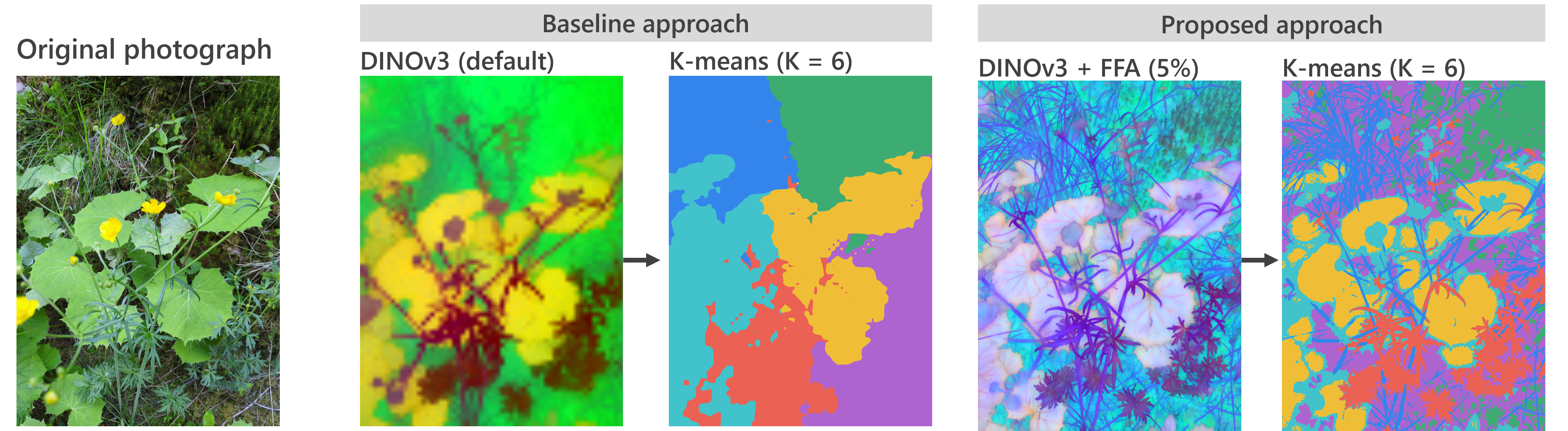}

{\footnotesize\textbf{Graphical summary.} Default DINOv3 and FFA (5\%) views of one vegetation photograph; K-means colors illustrate appearance groups, not biological labels.}
\end{center}

\begin{abstract}

Structures developed in natural systems are complex, and therefore, capturing the complexity through computer vision remains as a challenge. To address this, this study presents the Forced Focused Attention (\FFA) algorithm and its Python implementation, \texttt{vividRGB}. FFA is an encoder-agnostic algorithm for retrieving dense, various visual cues from RGB photographs without retraining a vision encoder. The procedure restricts input context to overlapping local crops, enlarges them, averages inverse-aligned reflections, and reconstructs a feature field in source coordinates. As a use case, FFA was applied to high-resolution image datasets relevant to ecology and evolution: grassland, fungal networks, airborne forest imagery, coral reefs, and phytoplankton microscopy. As a result, with a frozen DINOv3 ViT-B/16 encoder, \FFA\ yields higher mean RGB-edge boundary F1 than DINOv3 original and automatic SAM3 in each of five tested domains. The domain means range from 0.244 to 0.596 across DINOv3-FFA5\% and DINOv3-FFA10\%, compared with 0.028--0.076 for DINOv3 original and 0.020--0.090 for SAM3. The demonstrated capability of \FFA\ for edge and shape preservation suggests that generated feature maps can be used for downstream tasks which require preservation of visually dense structures.  \FFA's superiority is attributable to its rule to let the encoder ignore the broader contexts available in the whole image and forcefully focus on identifying variation at a much narrower scale. While \FFA\ performance generally increases by letting it focus on a narrower region, \FFA\ requires substantially greater extraction time and energy consumption than the case where the vision encoder is applied for the whole image. Therefore, this approach may be more useful for analyzing relatively small yet high-resolution image datasets which require dense, various visual cues. The alpha release \texttt{vividRGB} 0.1.0 and versioned research resources support reproducible cue inspection.

\end{abstract}


\section{Introduction}
\begin{figure}[H]
\centering
\includegraphics[width=\linewidth]{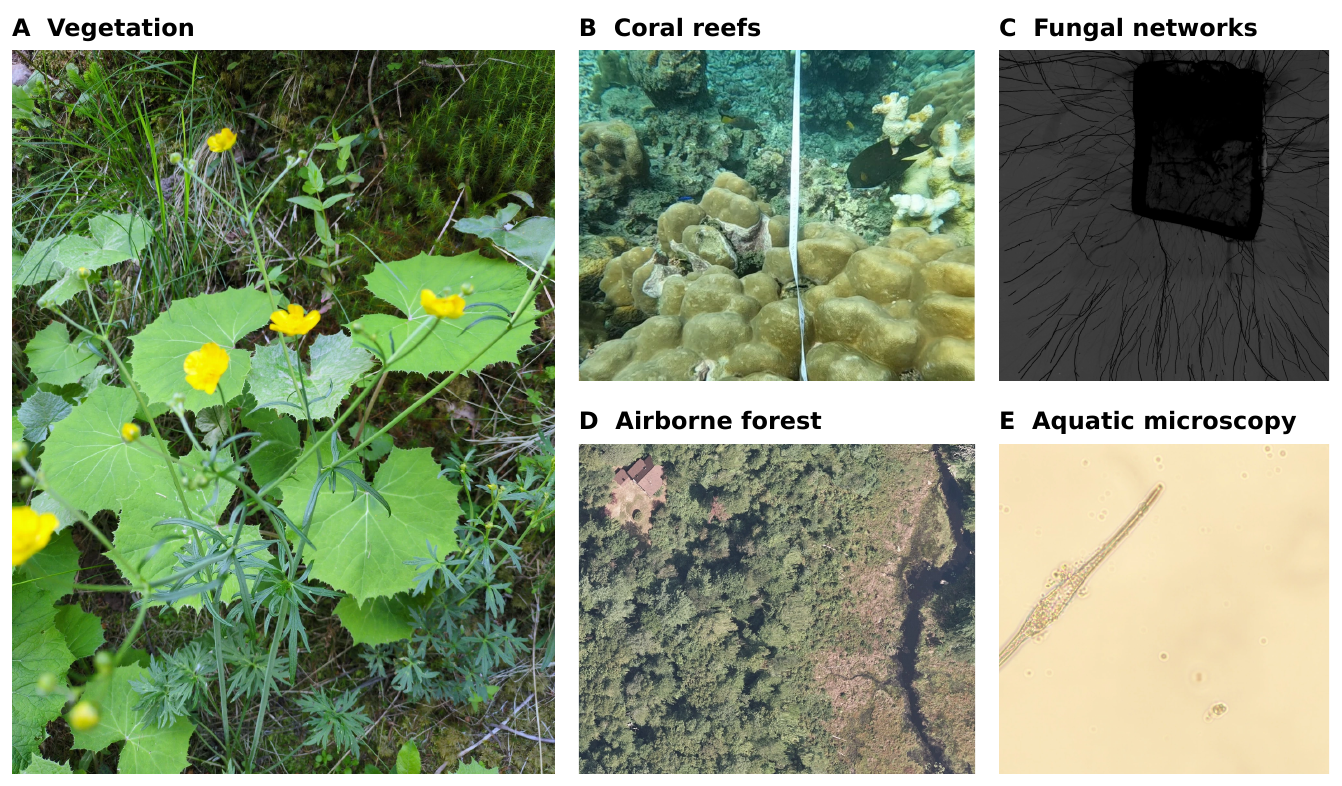}
\caption{Photographic examples across ecological scales. (A) Vegetation community in a project photograph, also used in the first row of Figures~\ref{fig:continuous} and~\ref{fig:qual}; (B) coral-reef scenery (central crop; source: \citep{coralscapes}); (C) fungal hyphae network (source: \citep{fungaldata}); (D) airborne forest imagery (source: \citep{neondata}); and (E) aquatic phytoplankton microscopy (source: \citep{pmid}). Panel sizes do not represent a common physical scale.}
\label{figure 1}
\end{figure}

Patterns found in nature reveal complex system properties such as hierarchically nested structure with patterns emerging from self organization \citep{simon1962,levin1998, riva2023}. From an ecological and evolutionary perspective, the patterns represent variation within and among organisms, populations, communities, and ecosystems \citep{levin1992,lewontin1970}. Structural patterns, for instance, can be morphological trait of plant leaves, branching architecture in network, colony structure in a community, heterogeneity within the population, and the juxtaposition of organisms and substrate  (Figure~\ref{figure 1}). Ecological structures often indicate underlying ecological processes and functions \citep{lavorel2002,noss1990}. Therefore, structural pattern monitoring and analysis serve as the basis for understanding the living systems on the planet. 

For monitoring and analyzing structural patterns, computer vision is a straightforward approach with imaging technologies. Recent vision foundation models are exemplified with DINO (Self-Distillation with No Labels) \citep{caron2021} and SAM (Segment Anything Model) \citep{kirillov2023} series. DINOv3 supplies self-supervised dense patch representations \citep{dinov3}; SAM3 supplies promptable segmentation and tracking \citep{sam3}. Both use Vision Transformers (ViT) as core parts of their architectures, dividing an image into patches and exchange information between tokens \citep{vit2021}. A patch descriptor can incorporate surrounding context through token interactions. Given their extensive coverage of image datasets, the contextual access supports representation learning and feature representations for various image types. In principle, these models should work well for detecting ecologically-relevant visual cues and structural patterns represented in (Figure~\ref{figure 1}). However, in practice, the story is different. As demonstrated in this paper, these foundation models reveal unsatisfactory results (yet) if they are used directly. 

A main reason why those vision foundation models cannot capture ecological structural complexity from images is that a single image contains too diverse, dense visual cues. Those foundation models are rather suited for identifying a few representative architectures in images. Diverse, dense structures could have been already captured in representation learning, but they are somewhat not reconstructed for feature mapping.  For instance,  (Figure~\ref{figure 1}) panel A shows a mixture of several plant species with visual cues such as yellow flowers, large round/heart-shaped leaves, grasses, edgy-shaped leaves, and moss. Objects overlap in multiple layers and partially occlude one another, fragmenting their visible shapes into disconnected regions.  DINOv3 cannot differentiate these items clearly. Meanwhile, SAM3 with text concepts, image exemplars, or their combination could address but remain challenging \citep{sam3}: Text concepts are hard to describe when objects are hard to name, and visual exemplars can easily get confused to which extent the object mask should cover because of the nested structures and too many objects. In short, those models are suited to capture relatively large-scale representative objects from the whole image, but they are not meant for capturing fine-scale representations of diverse, dense objects. 

In contrast, a human ecological observer can alternate between a community-level view and close inspection of a leaf, branch, or organism. Human visual-attention experiments likewise distinguish broader and narrower spatial allocation \citep{eriksen1986}. Visual cognition can draw on both global structure and local detail: experiments have demonstrated global precedence in hierarchical figures \citep{navon1977} and coarse-to-fine processing in rapid scene recognition \citep{schyns1994}. These findings provide evidence for processing at multiple scales within the tasks studied, without establishing a universal order of processing. Inspired by this multiscale capacity, this study introduces \emph{forced focused attention} (\FFA), an inference-time procedure that restricts the spatial context supplied to the vision encoder and integrates the resulting local representations across the full image.

FFA is a mode-agnostic algorithm with any encoder-decoder structure (but mainly for ViT). Technically, it presents enlarged, overlapping local crops to a frozen encoder, then align and combine their descriptors in source coordinates. "Attention'' denotes the imposed field of view, not a new attention layer, learned mask, or manipulation of attention coefficients. The encoder parameters remain unchanged. Reconstructing local descriptors permits inspection of fine cues together with their distribution across the scene.

The study investigates the importance of local focus on (high-resolution) images to preserve the details in the whole image patterns, which is critical for analyzing ecologically-complex visible structures. The results show that FFA maintains strong visible RGB edges and shapes successfully when feature maps are constructed across five ecological RGB image domains (Section~\ref{sec:datasets}). The evaluation distinguishes RGB-edge boundary F1 as class-boundary reference as well as downstream tasks like feature map visualization with principal component analysis (PCA) and clustering/segmentation.  The primary use case employs DINOv3, with a few additional examples using SAM3 (Supplementary Section~\ref{app:sam3-ffa}). Since computational power and energy consumption can increase exponentially with FFA approach, the analysis also investigates one key hyperparameter: to which narrow extent a focus should be given. 

The released Python \texttt{vividRGB} library (version 0.1.0) is available at \url{https://pypi.org/project/vividRGB/0.1.0/} \citep{vividrgb}.

\section{Related approaches and scope}
Related work situates inference scale, representation learning, reconstruction, and clustering as separate interventions.

\subsection{Dense visual descriptors and spatial sampling}
Amir et al. demonstrate dense DINO-ViT descriptors for correspondence, co-segmentation, and part discovery \citep{amir2021}. Their inference procedure increases descriptor sampling density by reducing the patch-embedding stride, allowing overlapping patches while retaining image-wide transformer context. Their supplement also uses flips and 95\% random-crop augmentations; cropping and augmentation are therefore not absent from this precedent. \FFA\ instead repeatedly encodes much smaller, enlarged fields of view and reconstructs their descriptors in source coordinates. The distinction concerns sampling and supplied context, not the first use of dense ViT features or crops.

Conceptually, FFA shares the hint of local observation with VICRegL. VICRegL learns global and spatially matched local representations during self-supervised training \citep{vicregl}. It addresses representation learning itself, whereas the present intervention changes the inputs and reconstruction of an already-trained encoder. Neither approach establishes that local training is universally necessary, or that inference-time cropping recovers biologically meaningful structures without validation.

\subsection{High-resolution inference with scales, tiles, and transformed views}
Scaling on Scales ($S^2$) evaluates a frozen encoder at several image scales, divides enlarged inputs into regular-sized subimages, reassembles and pools their features, and concatenates representations across scales \citep{s2}. Its parameter-free extraction wrapper is distinct from its downstream-head training and separate pretraining experiments. This establishes multiscale tiled inference as prior work; \FFA\ retains one focus at a time and evaluates reconstructed dense fields rather than claiming novelty for scale aggregation.

Docherty et al. transform inputs to a frozen DINOv2 encoder, upsample feature grids, inverse-align the transformations, and average the resulting maps \citep{docherty2026}. Their method uses shifts and flips, supports rotations, and is compatible with reduced patch-embedding stride. This is a direct precedent for the inverse-aligned reflection ensemble used here; the procedure appeared in their 2024 preprint before its 2026 journal publication. \FFA\ additionally specifies restricted enlarged crop contexts, a regular overlapping grid, and Hann-weighted reconstruction for the present ecological study. I do not claim that transformation averaging itself is new.

\subsection{Feature-map upsampling and hybrid reconstruction}
Feature-map upsampling estimates a denser field from coarse descriptors, while high-resolution inference obtains additional descriptors by re-encoding RGB at different scales or locations. FeatUp uses learned image-guided upsampling or per-image implicit fitting with multiview consistency \citep{featup}. FeatSharp combines upsampled global features with independently encoded local tiles through learned fusion \citep{featsharp}. It is consequently a close tiled-reconstruction precedent, not merely interpolation of a single feature map.

LoftUp learns coordinate-based high-resolution feature prediction using RGB guidance and cross-attention to coarse features; its later training stage uses crop outputs from a learned teacher upsampler as targets \citep{loftup}. AnyUp trains a universal, encoder-agnostic feature upsampler against crop-aligned descriptors from a frozen encoder \citep{anyup}. Their learned reconstruction components distinguish these methods from the retained \FFA\ procedure, which re-encodes enlarged crops and uses bilinear placement with normalized taper weights. None of these learned upsamplers, including AnyUp, is part of the retained pipeline. These methods have not been benchmarked on the five-domain collection, so the present results cannot establish superiority over these reconstruction methods.

\subsection{Ecological crop inference and the scope of FFA}
The verified Foveate workshop record describes exemplar-guided instance segmentation with frozen DINOv3: exemplar similarity gates foreground patches, connected components propose tighter crops, and recursive re-embedding refines instances \citep{foveate}. This supports ecological crop inference as relevant prior work. Its abstract describes adaptive, target-guided zooming, whereas \FFA\ uses declared regular crop geometry for image-only cue reconstruction. Only the record and abstract were verified; they do not establish whether Foveate also uses reflections, overlap tapering, or other reconstruction details.

The contribution of this study is the specified ecological workflow and its evidence: local crop enlargement and context restriction, inverse-aligned reflection averaging, overlap-weighted reconstruction, appearance-boundary evaluation, and analysis of clustering complexity and computational cost. The individual ingredients have precedents. DINOv3 features are clustered, while automatic SAM3 masks are resolved by a declared overlap rule; neither human text nor exemplar prompts are supplied to SAM3 \citep{hfsam3}. This common RGB-edge endpoint does not make their outputs semantic equivalents or their computation equal. Methods specifies the controls, and Discussion distinguishes exploratory cues from biological segmentation.

Grouping is a separate summary stage: silhouette and BIC assess a feature partition or density model, not taxonomic identity \citep{rousseeuw1987,schwarz1978}. Varimax rotates a retained subspace for inspection \citep{kaiser1958}. The tested complexity penalty and display choices therefore require their own interpretation alongside feature reconstruction.

\noindent\begin{minipage}{\linewidth}
\vspace{\baselineskip}\section{Materials and methods}

\subsection{\emph{Forced Focused Attention} (FFA) algorithm workflow}
FFA combines local inspection with global aggregation. When a Vision Transformer (ViT) processes an entire image, its self-attention can relate local features to the broader scene \citep{vit2021}. FFA instead restricts each encoder input to a small crop, excluding the surrounding image from that inference pass. The term ``forced focused'' describes this externally imposed restriction of the model's visual context. Enlarging the crop makes local edges, shapes, textures, and colour differences more prominent within the encoder input. Features extracted from overlapping crops are then blended back into a whole-image representation. The approach changes the input context without retraining the encoder or changing its learned parameters.
\end{minipage}\par\smallskip
\begin{enumerate}[leftmargin=*,itemsep=3pt,topsep=3pt]
\item \textbf{Local focus for feature extraction.} First, choose a focus fraction $f$ relative to the working image's short side $L=\min(W,H)$, giving a square crop side $c_f=\max\{c_{\min},\operatorname{round}(fL)\}$. For instance, $f=0.10$ specifies 10\% of the short-side length of the image, not 10\% of the image area. Second, slide the crop along both axes with a step of approximately $(1-\alpha)c_f$, where $\alpha$ is the overlap fraction ($\alpha=0$ means no overlap), and add edge-aligned windows to cover the entire image. Third, resize each crop by bicubic interpolation to the chosen encoder input resolution, and apply the checkpoint's RGB rescaling and channel standardization. Fourth, extract spatial patch descriptors with a pretrained encoder such as DINOv3. Following Docherty et al.\ \citep{docherty2026}, token grids from the original, horizontally reflected, vertically reflected, and doubly reflected versions of each rescaled crop are inverse-aligned and averaged. For the DINOv3 comparison in Table~\ref{tab:f1}, the FFA settings were $f=0.05$ or $0.10$, $c_{\min}=16$ pixels, nominal $\alpha=0.75$, and an encoder input resolution of $384\times384$ pixels; these are experimental choices rather than requirements of FFA.
\item \textbf{Global aggregation from locally extracted features.} Bilinearly interpolate each local feature map to its source crop's pixel grid and place it back at the corresponding coordinates in the working image. Where crops overlap, weight their descriptor vectors with a floored separable Hann taper, sum the weighted vectors, and divide by the total coverage weight. A Hann taper (also called a Hann window or cosine bell) is a bell-shaped weighting function that emphasizes crop centres and downweights their edges, helping reduce stitching artefacts when overlapping feature maps are blended \citep{pielawski2020}. This normalized aggregation produces one image-aligned continuous feature field for each focus setting; it averages features, not display colours or cluster identifiers.
\end{enumerate}

\subsection{Datasets for FFA experimentation}\label{sec:datasets}
The datasets were selected to expose FFA to varied ecological settings, organism groups, and scales of visible structure. They span microscopic phytoplankton \citep{pmid} and fine fungal networks \citep{fungaldata}, ground-level grassland vegetation \citep{kicszert}, underwater reef communities \citep{coralscapes}, and airborne views of forest canopies \citep{neondata} (Fig.~\ref{figure 1}). Together, they include fungi, terrestrial plants, aquatic microorganisms, and reef organisms, with contrasting branching networks, cell shapes, overlapping leaves, benthic mosaics, and canopy arrangements. This range supports exploratory assessment of local-focus reconstruction from fine organismal detail to broader community structure. The fungal source includes both microscopic and macroscopic laboratory-grown mycelial networks, including growth on agar and compressed soil \citep{aguilar2022}; the collection therefore spans observation settings as well as ecological subjects.

The analysis set contains 50 working RGB images, ten per domain (Table~\ref{tab:data}); these same images are reused for comparisons of methods, clustering analyses, and tests of crop size under their separately declared protocols. The working images are derived from source photographs for inference and evaluation, without training or fine-tuning the encoders on this collection. The source collections and selection rules are:

\begin{itemize}[leftmargin=*,itemsep=3pt,topsep=3pt]
\item \textbf{Grassland vegetation: MOIN/KICS-Zert.} Project photographs of mixed vegetation at four German sites are associated with the KICS-Zert biodiversity-monitoring project \citep{kicszert}. The ten MOIN examples cover earlier within-site visual-rank strata: high- and low-rank-quintile photographs from each site, supplemented by middle-quintile examples at Fe\ss bach and Heidesheim.
\item \textbf{Fungal networks: Fungal Network Analysis.} The Zenodo collection provides branching mycelial-network photographs \citep{fungaldata,aguilar2022}. Selection is stratified by the source folders: four ascomycota, three basidiomycota, and three zygomycetous images.
\item \textbf{Airborne forest imagery: NeonTreeEvaluation.} Archived NEON RGB canopy tiles are sampled from the benchmark's training split, one per selected site, with source short side at least 1024 pixels \citep{neondata}. This study uses these tiles only for inference and evaluation. The imagery is airborne, rather than UAV imagery.
\item \textbf{Coral reefs: Coralscapes.} Benthic-community frames and their semantic class labels are obtained from Coralscapes \citep{coralscapes}. Selection first samples a test frame per available site and then fills the remaining slots. The working images retain the distributed $2048\times1024$ frame dimensions.
\item \textbf{Phytoplankton microscopy: PMID2019.} Laboratory microscopy images are selected from 47 image--annotation pairs in the accessible initial public-download listing, rather than from the entire 10,819-image dataset \citep{pmid}. They supply an additional morphological setting; microscopy does not necessarily depict a dense multispecies community.
\end{itemize}

For the five-domain analyses, EXIF orientation is applied when specified, and a native working window is taken without enlarging the source image. Working dimensions are compatible with the encoder's 16-pixel patch grid; source and working dimensions, window coordinates, and SHA-256 hashes are recorded per image. Large fungal originals, up to approximately $27{,}000\times18{,}000$ pixels, and NEON tiles, up to $10{,}000\times10{,}000$, were cropped to central $2048\times2048$-pixel working windows rather than downsampled in full. For these sources, each window dimension was capped at 2048 pixels and rounded down to a multiple of 16, permitting smaller or nonsquare windows when required by the original dimensions. Only pixels within the retained window were analyzed; the source pixels were not interpolated or enlarged. Resizing individual FFA crops to the encoder input resolution was a separate, later preprocessing step. 

\begin{table}[tb]
\centering\small\setlength{\tabcolsep}{3pt}
\caption{The selected five-domain image collection. Dimensions are those actually supplied to this experiment, expressed as width $\times$ height; they are not nominal dimensions of every source image. Each domain contributes $n=10$.}
\label{tab:data}
\begin{tabular}{>{\raggedright\arraybackslash}p{31mm}>{\raggedright\arraybackslash}p{49mm}>{\raggedright\arraybackslash}p{37mm}}
\toprule
Domain & Source and content & Working dimensions (pixels) \\
\midrule
Grassland & MOIN/KICS-Zert; mixed vegetation at four German sites & $2048\times2048$ \\
Fungal networks & Zenodo 5725751; branching mycelial networks \citep{fungaldata,aguilar2022} & $2048\times2048$ \\
Airborne forest & NeonTreeEvaluation; NEON RGB canopy imagery \citep{neondata} & $1312\times1216$ to $2048\times2048$, including nonsquare windows \\
Coral reefs & Coralscapes; benthic community frames \citep{coralscapes} & $2048\times1024$ \\
Phytoplankton & PMID2019; laboratory microscopy \citep{pmid} & $1360\times1024$ or $2048\times2048$ \\
\bottomrule
\end{tabular}
\end{table}

\subsection{Evaluation}\label{sec:analysis-design}\label{sec:evaluation}
Evaluation focuses on whether local-focus reconstruction improves the preservation of visible image boundaries, how that outcome depends on focus fraction $f$, and what computational cost it incurs. FFA was applied with different settings to the collection of 50 working images (Section~\ref{sec:datasets}). RGB-edge recovery was the primary quantitative endpoint in this study; visual displays and sensitivity analyses of automatic grouping provide separate checks. Further evaluation could address downstream tasks such as prediction, object detection, and counting. Here, the evaluation focuses on appearance-boundary recovery and computational cost.

\subsubsection{Models and baseline comparisons}\label{sec:eval-models}
The primary feature comparison uses a frozen DINOv3 ViT-B/16 encoder \citep{dinov3}. The \emph{DINOv3 original} control processes the complete working image once at its native working resolution and in one orientation. The FFA5\% and FFA10\% conditions instead use DINOv3 to encode overlapping local crops and reconstruct their descriptors on the same working-image grid. The 5\% and 10\% values of $f$ were chosen subjectively; a wider range was tested in the crop-size analysis (Section~\ref{sec:eval-focus}). The backbone weights are identical across conditions. DINOv3 supplies 768-dimensional patch descriptors at a 16-pixel stride; local crops are rescaled to $384\times384$ pixels before encoding. ``Original'' identifies this control protocol, rather than the uncropped source photograph or the checkpoint's default thumbnail size.

In addition to DINOv3, an \emph{automatic SAM3 mask-generation pipeline} was tested as a baseline \citep{sam3,hfsam3}. In this study, the pipeline takes the image as input and internally generates a $32\times32$ point-prompt grid, without manually selected prompts. Overlaps are resolved using predicted mask IoU scores; uncovered pixels remain background. Its boundaries come from the resulting mask partition, whereas DINOv3 boundaries come from clustered encoder features. Model revisions, preprocessing, mask filters, and numerical settings are recorded in Supplementary Section~\ref{app:encoder-settings}.

\subsubsection{From feature maps to predicted boundaries}\label{sec:eval-feature-boundaries}
PCA-based visualization and clustering were employed as downstream applications of the reconstructed feature field. Each DINOv3 patch descriptor contains 768 learned feature values. Principal component analysis (PCA) identifies orthogonal directions of decreasing variance \citep{jolliffe2016}; here, each descriptor is represented by 16 scores along the leading directions estimated from a calibration sample. Sixteen is a fixed analysis setting, rather than a dimension count selected by an explained-variance threshold. For each image in Table~\ref{tab:f1}, a common PCA mean and basis are estimated from balanced samples across conditions and reused for all treatment fields. Mean centering is applied without whitening, so component variances are not forced to be equal.

Each treatment is independently partitioned using K-means \citep{lloyd1982} with $K=6$, fitted to the 16-score feature vectors at the same sampled image positions. The position indices locate the vectors; image coordinates are not added as clustering features. Each model estimates its own six cluster centers, and all source-grid locations are then assigned to their nearest center using distances in the 16-dimensional PCA space. A pixel becomes a predicted boundary when its label differs from its right or lower neighbor. This evaluates the encoder--partition combination without requiring arbitrary cluster IDs to correspond across methods. The choice of $K=6$ was subjective, and $K=12$ was also tested in a sensitivity analysis (Supplementary Section~\ref{app:fixed-partitions}); the supplement also records sampling and fitting settings.

For visualization, the first three ordinary PCA scores, or the first three axes of the full 16-dimensional varimax rotation \citep{kaiser1958}, supply red, green, and blue display channels. Stretching and clipping are applied only to these false-color displays; primary clustering uses all 16 scores before display transformations (Supplementary Section~\ref{sec:display-methods}).

For exploratory visual grouping, the number of clusters is also selected automatically rather than fixed at six. Automatic K-means compares $K=3,\ldots,12$ using held-out silhouette scores, favoring counts closest to five among candidates within one standard error of the best score. Full-covariance Gaussian mixture models compare $K=1,\ldots,12$ using a BIC-style fit--complexity criterion with a strengthened penalty; the displayed maps use a multiplier of four, chosen after inspecting these images. These choices summarize appearance variation and do not estimate a known number of biological groups. Supplementary Section~\ref{sec:group-methods} gives the selection procedures and their sensitivity to the penalty and retained dimension.

\subsubsection{Reference boundaries and performance metrics}\label{sec:eval-reference-metrics}
The primary evaluation asks whether reconstructed boundaries follow strong visible changes in an image. Applying the same image-derived reference rule across the five domains permits this comparison without requiring reviewed object-boundary annotations in every dataset. Each unclustered RGB image is converted to grayscale brightness and lightly smoothed to reduce pixel-scale fluctuations; strong local brightness changes then define the reference edges, independently of the encoder, PCA, or clustering. These appearance edges can include texture, shadows, and imaging artifacts, while missing weak transitions or color boundaries with similar brightness. They are not verified biological object outlines.

Predicted and reference edges are restricted to a valid interior obtained by eroding the entire working-image domain by two pixels. With a two-pixel Manhattan tolerance, precision measures the fraction of predicted boundary pixels near a reference edge, and recall measures the fraction of reference pixels reached by a predicted boundary. Their harmonic mean is RGB-edge boundary F1 (Eq.~\ref{eq:f1}). This tolerant set comparison does not measure region IoU or biological segmentation accuracy. Boundary density is reported because additional boundaries can improve recall. 

For comparison with the FFA-based results, the corresponding RGB images were also partitioned directly using the same K-means procedure. This control tests how directly partitioning RGB appearance recovers the selected visible edges. Direct RGB clustering was expected to perform well on this appearance-based endpoint. Higher RGB-edge boundary F1 indicates better agreement with the RGB-derived edge reference. Performance was evaluated separately for each image and summarized by ecological domain. The model comparisons in Table~\ref{tab:f1} used paired scores from the same images; the paired-comparison procedures and limitations are given in Supplementary Section~\ref{app:contrasts}.

\subsubsection{Effect of local-focus size}\label{sec:eval-focus}
The focus fraction $f$ determines how crop scale and supplied context affect recovery of RGB appearance boundaries. Alongside the comparison at 5\% and 10\% focus, the range was tested at eight complete settings: 100\%, 40\%, 20\%, 10\%, 5\%, 4\%, 3\%, and 2\% of the working image's short side. Each setting covers all 50 images, giving 400 image--focus observations in total ($50\times8$). The control is native-resolution, single-orientation DINOv3 original. All crop-size settings use approximately 75\% overlap and coverage-normalized reconstruction. RGB-edge boundary F1 was expected to increase as crop size decreased with lower $f$. Crop geometry and measurement-stage definitions are given in Supplementary Sections~\ref{app:reconstruction-details} and~\ref{app:focus-cost-details}.

\subsubsection{Computational cost and performance--energy tradeoff}\label{sec:eval-cost}
Energy consumption was expected to increase as crop size decreased with lower $f$. To test this, the operational cost of the declared inference stages was quantified on one NVIDIA RTX 4000 SFF Ada GPU. DINOv3 feature extraction for the crop-size analysis included encoding and two PCA projection/reconstruction calculations. It excluded calibration, checkpoint I/O, power-bracketing waits, and clustering/scoring. Board power was sampled approximately every 200 ms. Energy was estimated by integrating board power over the measured stage without idle-baseline subtraction; it excluded CPU and host-memory energy, wall-plug losses, and carbon accounting. Efficiency was calculated as inverse joules per image, summarized as the mean of individual reciprocals. Detailed timing boundaries and the integration formula are in Supplementary Section~\ref{app:focus-cost-details}.

Across the crop-size range, both RGB-edge boundary F1 and energy consumption were expected to increase as $f$ decreased. F1 is bounded by 1, whereas finer crops can require more energy. The analysis aimed to identify a suitable balance between gains in F1 and energy consumption. For adjacent complete focus settings, paired changes in RGB-edge boundary F1 and energy were averaged within domains and then equally across domains. Marginal summaries express F1 gain per additional 1000 J and joules per $+0.01$ F1 score. Exploratory analyses of crop-size choice compare continuous hinge fits across coordinates and focus ranges, and evaluate observed mean F1 after subtracting application-specified penalties on mean energy. Procedures and interpretation limits are given in Supplementary Sections~\ref{app:focus-cost-details} and~\ref{app:operating-point}.

For an adjacent coarser-to-finer step, let $\Delta E=\frac15\sum_{d=1}^{5}\operatorname{mean}_{i\in d}(E_{i,f}-E_{i,c})$ and $\Delta F=\frac15\sum_{d=1}^{5}\operatorname{mean}_{i\in d}(F_{i,f}-F_{i,c})$, using the same paired images for energy and F1. Marginal gain and incremental cost are summarized as
\begin{equation}
G_{1000}=1000\frac{\Delta F}{\Delta E},\qquad C_{0.01}=0.01\frac{\Delta E}{\Delta F}.
\label{eq:marginal-energy}
\end{equation}
$G_{1000}$ gives F1 gain per extra 1000 J when $\Delta E>0$; $C_{0.01}$ gives joules per $+0.01$ absolute F1 when both differences are positive. The latter scales the observed step and does not predict that another 0.01 gain is achievable. Ratios are calculated from paired means, not by averaging unstable individual ratios. Intervals follow the stratified paired-image bootstrap, including its rule for withholding ratios whose denominator interval reaches zero. 

\subsection{Implementation and reproducibility}\label{sec:implementation}
Crop geometry, reflection alignment, bilinear interpolation, and coverage-normalized Hann weighting are specified in Supplementary Section~\ref{app:reconstruction-details}. Supplementary Section~\ref{app:evaluation-details} records checkpoint revisions, partition-fitting settings, boundary scoring, and display transformations; Supplementary Section~\ref{app:focus-cost-details} defines the measurement stages, and Supplementary Table~\ref{tab:parameters} summarizes selected experimental settings. These records distinguish the saved results comparing methods from later analyses and support saved-result verification without substituting settings between experiments.

\FloatBarrier

\clearpage
\section{Results}

\subsection{Visual inspection: PCA and grouping}
\subsubsection{PCA} 
\begin{figure}[p]
\centering\includegraphics[width=\linewidth,height=0.76\textheight,keepaspectratio]{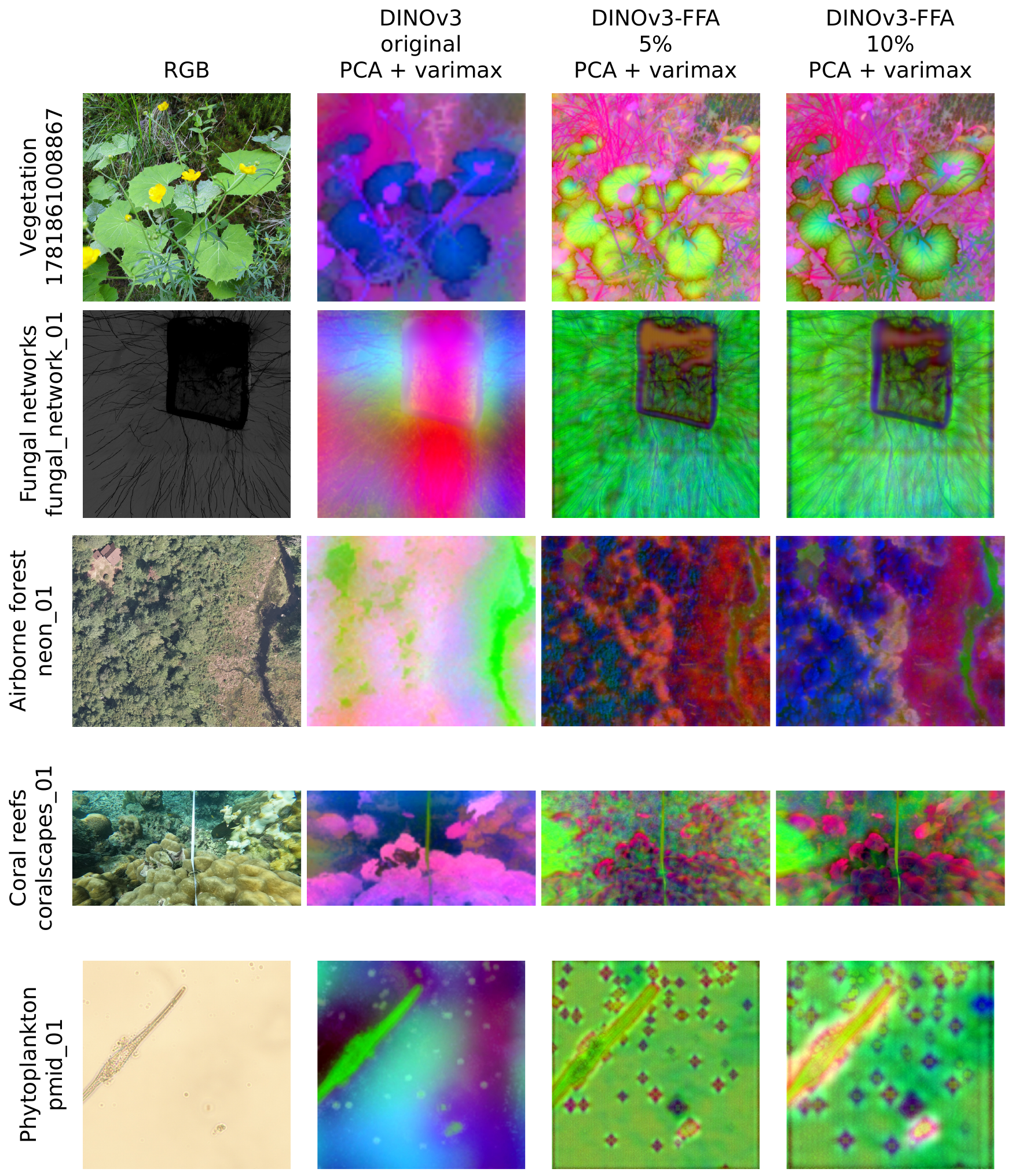}
\caption{RGB inputs and PCA--varimax false-color displays for five illustrative images. The first row uses the same square crop of the vegetation photograph in Figure~\ref{figure 1}A (photo \texttt{1781861008867}) in every column, as an additional illustration outside the 50-image benchmark; the other four rows retain benchmark examples. Columns show RGB, DINOv3 original, DINOv3-FFA5\%, and DINOv3-FFA10\%. Each image has a centered PCA16 basis and orthogonal 16-dimensional varimax rotation shared across its DINOv3 treatments; the first three rotated axes supply display colors. The vegetation row pools equal score samples from the three displayed treatments for its shared 2nd--98th percentile limits. The other rows retain archived limits that also include FFA20\%. Colors are feature coordinates, not biological classes. Other image sources: fungal Zenodo 5725751 \citep{fungaldata}; NEON/NeonTreeEvaluation \citep{neondata}; Coralscapes \citep{coralscapes}; PMID2019 \citep{pmid}.}
\label{fig:continuous}
\end{figure}

Figure~\ref{fig:continuous} first compares the RGB inputs with DINOv3 without FFA, with FFA5\% and FFA10\%,  where the false color composites were generated with PCA--varimax (R,G,B corresponds to the values of PC1, 2, and 3 after scaling, respectively). When checking the outcomes of DINOv3 original display (second column), an visual observation suggests that the stuctures of coral reefs (4th row) and phytoplankton (5th row) are relatively well preserved in DINOv3, while those of the vegetation community, fungal hyphae network, and forest canopies (1st--3rd rows) are weakly preserved in the feature embeddings. This indicates that dense visual cues are relatively challenging to be kept in the model. 

In contrast, DINOv3 with FFA5\%/10\% (3rd/4th columns) preserve the structural details of single leaves in the vegetation (1st row), each hyphae string of the fungi (2nd), and individual canopies of the forest and water courses (3rd). This visual inspection supports the key idea of the study that letting a foundation model focuses only on narrow regions of the image instead of the whole extent can help identifying the contrastive differences among objects including edges and shapes densely distributed in narrow regions. Meanwhile, FFA may not necessarily be effective when images are rather homogeneous with lesser dense structures (e.g. 5th row; phytoplankton). 

The definitions of ordinary PCA and varimax displays are provided in Supplementary Section~\ref{sec:display-methods}.

\subsubsection{Grouping} 
\begin{figure}[p]
\centering\includegraphics[width=\linewidth,height=0.79\textheight,keepaspectratio]{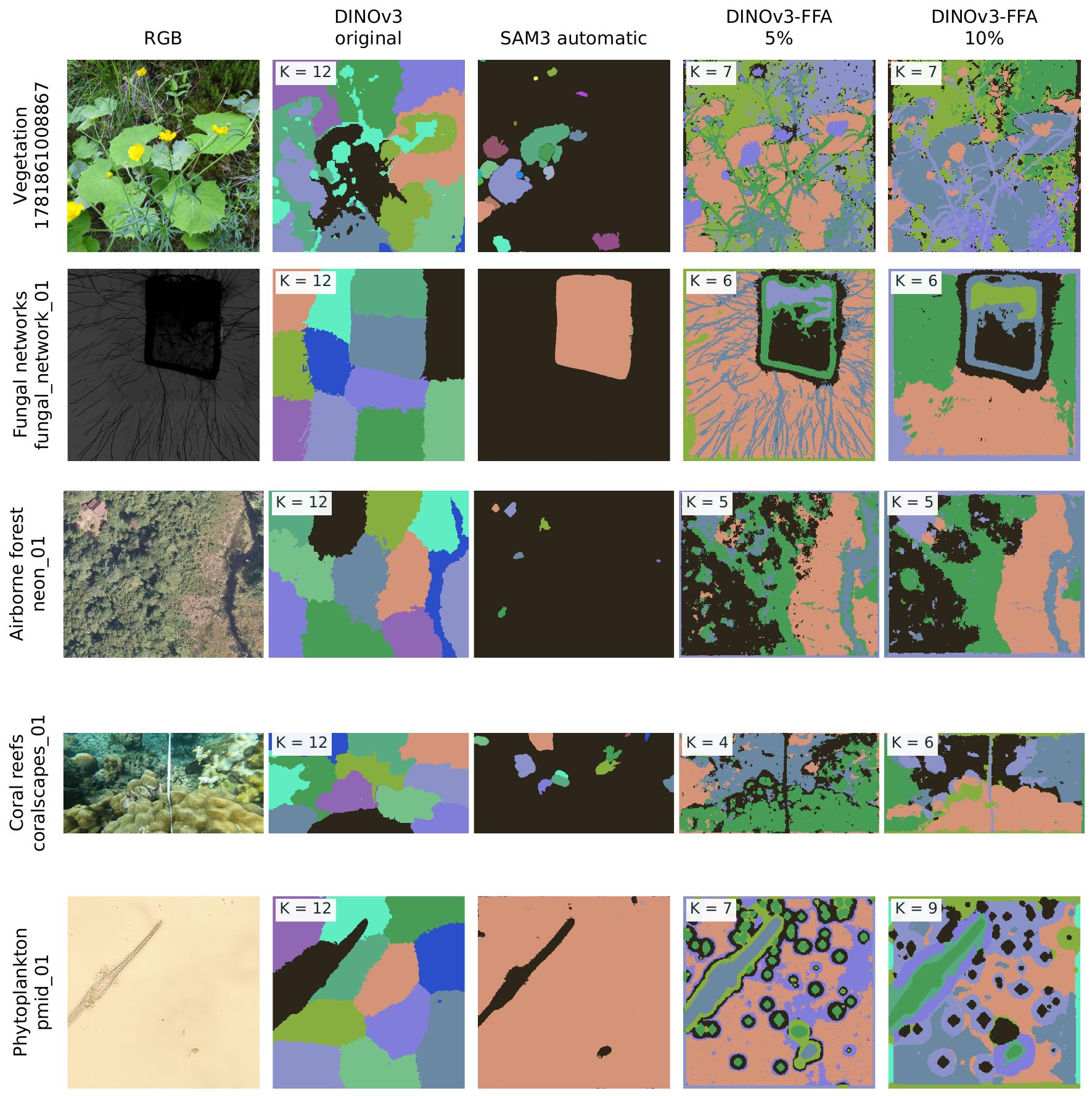}
\caption{RGB and appearance partitions for five illustrative images. The vegetation row uses the same square crop of the photograph in Figure~\ref{figure 1}A (photo \texttt{1781861008867}) in every column, as an additional illustration outside the 50-image benchmark; the other four rows retain benchmark examples. Columns show RGB, DINOv3 original, automatic SAM3, DINOv3-FFA5\%, and DINOv3-FFA10\%. The three DINOv3 treatments use independent full-covariance GMMs on their image-specific 16-dimensional PCA fields; labels show the selected tested component counts under the \(\lambda=4\) penalty. Automatic SAM3 retains its accepted masks and uncovered background. Colors are arbitrary partition labels and do not match across methods. Appearance partitions can reflect substrate, illumination, and optical artifacts; biological identities are not inferred. Other image sources: fungal Zenodo 5725751 \citep{fungaldata}; NEON/NeonTreeEvaluation \citep{neondata}; Coralscapes \citep{coralscapes}; PMID2019 \citep{pmid}.}
\label{fig:qual}
\end{figure}

Figure~\ref{fig:qual} shows the same set of example images with clustering outcome, as a potential downstream task. In addition to DINOv3, here SAM3 automatic detection is visualized (3rd column). It is noted that for clustering the 16 components were used, and therefore not necessarily express a visual association with the stuctures mapped with the top 3 components in Figure~\ref{fig:continuous}.  These displays use independent full-covariance GMM partitions with automatic component selection. For the sensitivity of automatic clustering to retaining three versus sixteen PCA dimensions, see Supplementary Table~\ref{tab:dimensions-condensed}. DINOv3 original selects $K=12$, the upper search bound for automatic search, in all five displayed examples; DINOv3-FFA5\% selects 4--7 components and DINOv3-FFA10\% selects 5--9. 

DINOv3 (2nd column) shows partitions rather randomly and evenly across the images without morphologically reasonable structuring (a few exceptions can be the river corridor in 3rd row and the phytoplankton in 5th row). In contrast, FFA-based partitions expose finer appearance changes: around vegetation leaves (1st row); network branches (2nd row); forest--grassland--river continuum (3rd row); reef structures (4th row); and microscopic forms with small particles (5th row). Automatic cluster-count selection is specified in Section~\ref{sec:eval-feature-boundaries}; its penalty and retained-dimension sensitivity results are reported in Supplementary Section~\ref{sec:group-methods}.

\FloatBarrier

\subsection{FFA preserves RGB-edge shapes: The Finer, the better?}
Across the same 50 images (ten per domain), the DINOv3 original, DINOv3-FFA10\%, and DINOv3-FFA5\% partitions, each fitted to 16 PCA coordinates, retain the mean F1 values from that calibration of 0.0479, 0.3164, and 0.4695, respectively. As the baseline, a mean F1 value was 0.6570 for direct RGB image clustering (\textit{K} = 6). DINOv3 with FFA had a substantially higher mean RGB-edge boundary F1 than DINOv3 original and automatic SAM3 in all five domains, and 5\% focus has higher means than 10\% focus in every domain (Table~\ref{tab:f1}). Holm adjustment yields $p=0.03906$. Supplementary Section~\ref{app:contrasts} supplies details. 

\begin{figure}[H]
\centering\includegraphics[width=\linewidth]{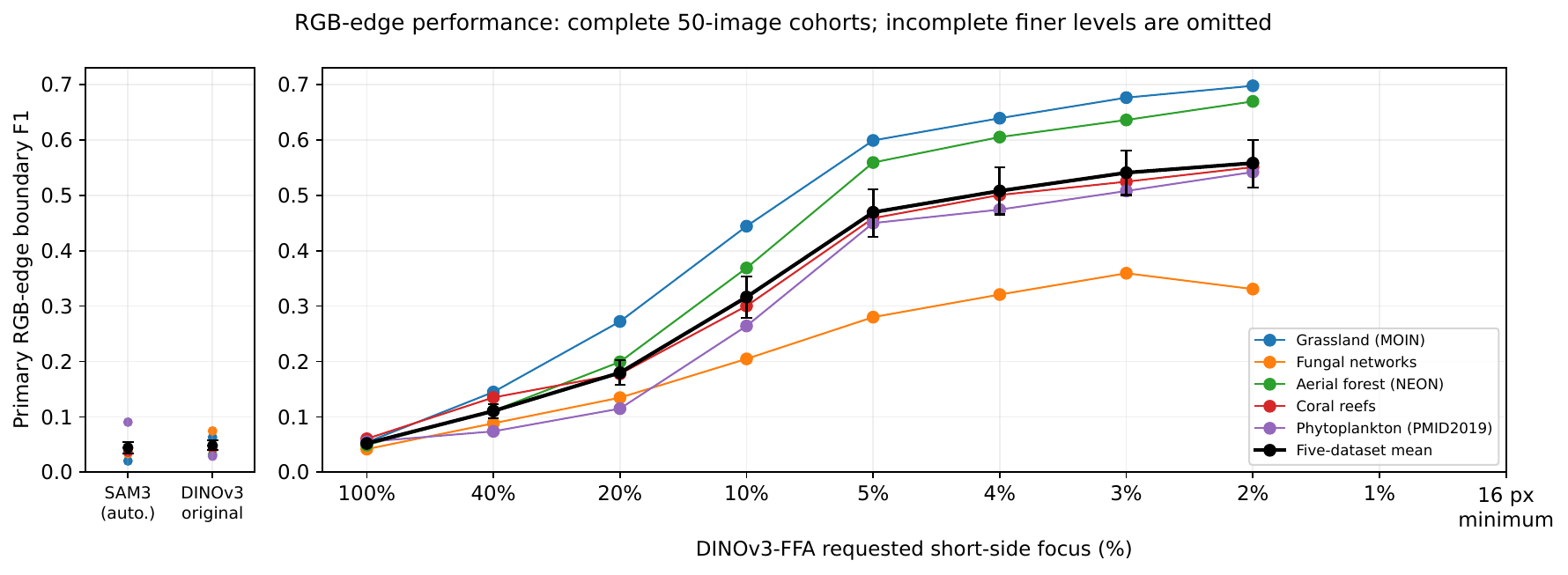}
\caption{Primary RGB-edge boundary F1 across crop sizes of \textit{f}. The left panel shows reused automatic SAM3 and DINOv3 original. Black points are equal-five-domain means with bootstrap 95\% 
intervals. The right panel shows DINOv3-FFA at different focus level of \textit{f}, with the same F1 scale. 100\% indicates almost equivalent to the whole image analysis as with DINOv3 original. Complete cohorts contain ten images per domain, 50 overall; curves end at 2\%.}
\label{fig:f1}\label{fig:gradient-primary}
\end{figure}

The figure~\ref{fig:f1} shows an increase in F1 score along with an increase in finer focus (i.e. lower \textit{f} value), yet the S-shaped curve indicates performance improvement hits a plateau at some moment. This suggests that local focus can help preserve the structural edges and shapes but the capability does not necessarily improve by focusing narrower regions.

\begin{table}[H]
\centering\small
\caption{Primary RGB-edge boundary F1 (mean $\pm$ sample SD; ten images per domain). The highest mean in each domain is bold. DINO-derived partitions use independent $K=6$ K-means. SAM3 uses automatic masks with highest-predicted-IoU overlap priority. }
\label{tab:f1}
\begin{tabular}{lcccc}
\toprule
Domain & \shortstack{DINOv3\\original} & SAM3 automatic & \shortstack{DINOv3-FFA\\5\%} & \shortstack{DINOv3-FFA\\10\%} \\
\midrule
% BEGIN GENERATED TABLE f1
Grassland (MOIN) & 0.061 $\pm$ 0.026 & 0.020 $\pm$ 0.016 & \textbf{0.596} $\pm$ 0.106 & 0.441 $\pm$ 0.083 \\
Fungal networks & 0.076 $\pm$ 0.057 & 0.034 $\pm$ 0.027 & \textbf{0.290} $\pm$ 0.108 & 0.244 $\pm$ 0.093 \\
Airborne forest (NEON) & 0.034 $\pm$ 0.011 & 0.041 $\pm$ 0.044 & \textbf{0.549} $\pm$ 0.136 & 0.360 $\pm$ 0.135 \\
Coral reefs (Coralscapes) & 0.041 $\pm$ 0.010 & 0.036 $\pm$ 0.024 & \textbf{0.436} $\pm$ 0.136 & 0.266 $\pm$ 0.074 \\
Phytoplankton (PMID2019) & 0.028 $\pm$ 0.007 & 0.090 $\pm$ 0.032 & \textbf{0.456} $\pm$ 0.118 & 0.288 $\pm$ 0.138 \\
% END GENERATED TABLE f1
\bottomrule
\end{tabular}
\end{table}

\FloatBarrier

\subsection{Finer focus needs a considerable computational cost}
\begin{figure}[!htbp]
\centering\includegraphics[width=\linewidth]{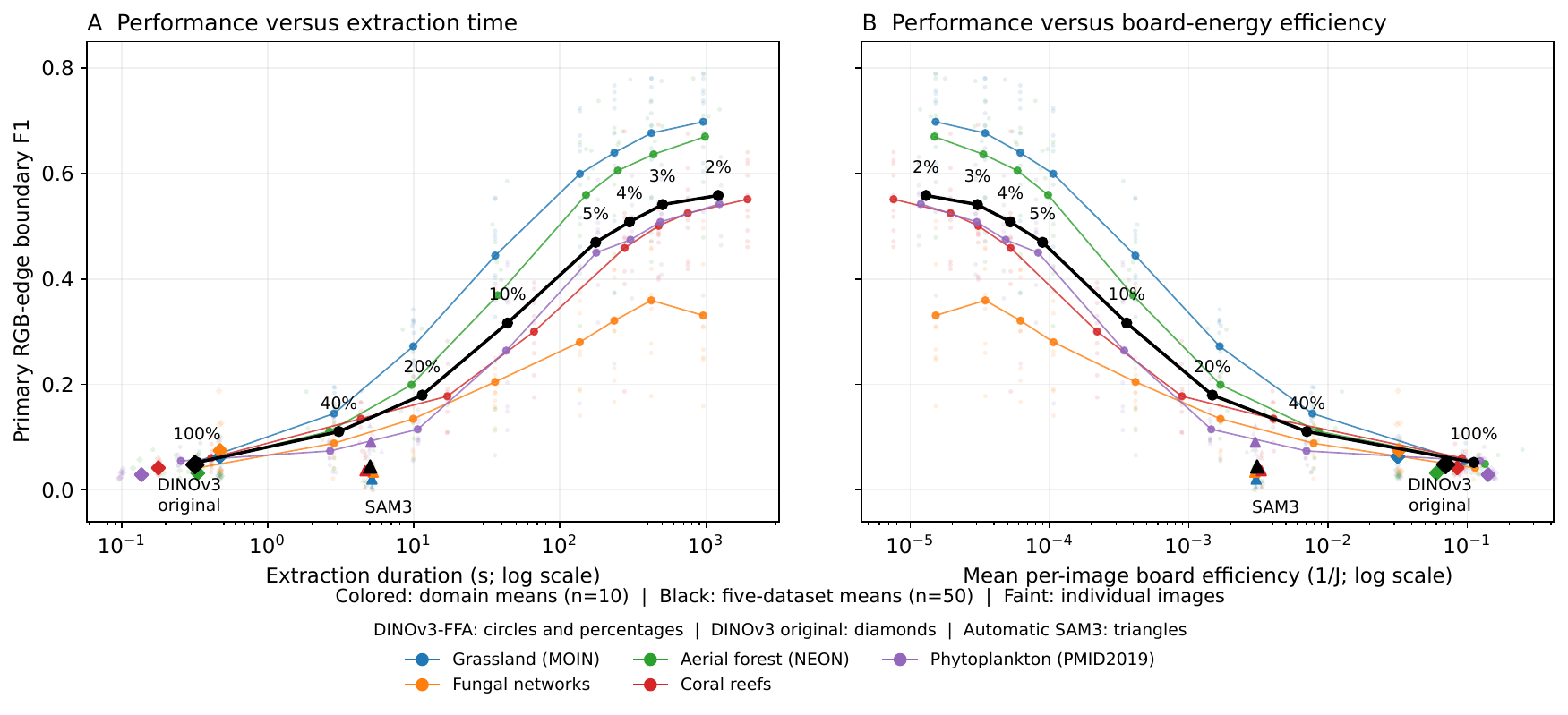}
\caption{Performance--computation trade-off across crop-sizes, also showing DINOv3 original and automatic SAM3. (A) Primary RGB-edge boundary F1 versus extraction duration. (B) F1 versus GPU board-energy efficiency, $1/J$. Horizontal axes are logarithmic; higher F1 with shorter duration or greater efficiency is preferable. Faint points show individual images, colored markers are domain means ($n=10$), and black markers are equal-five-domain means ($n=50$). Circles and percentage labels mark DINOv3-FFA; diamonds mark DINOv3 original and triangles SAM3.}
\label{fig:cost}
\end{figure}

While seeing an S-shaped curve with F1 score improvement (Figure ~\ref{fig:f1}), narrowing a focus inreases computational cost drastically. This concerns the balance between performance and cost. Table~\ref{tab:cost} and Figure~\ref{fig:cost} summarize DINOv3 original and DINOv3-FFA across crop sizes, with complete cohorts through 2\% focus. Mean extraction duration is just 0.317 s for DINOv3 original and 5.0 s for SAM automatic, while DINOv3 with FFA drastiacally slows down computation: 43.8 s at 10\%, 175.8 s at 5\%, and 1210.6 s at 2\% (20 minutes per image); mean GPU board energy is 2.96, 12.05, and 83.03 kJ, respectively, but the original DINOv3 and SAM require only 0.021 kJ and 0.32 kJ. 
Halving crop side approximately quadruples the number of overlapping windows, while every rescaled crop still requires four $384\times384$ encoder evaluations. 

\begin{table}[!htbp]
\centering\small
\caption{Single-run implementation cost for the newly measured DINOv3 original control and complete DINOv3-FFA cohorts at 100\%, 40\%, 20\%, 10\%, 5\%, 4\%, 3\%, and 2\%, plus reused automatic SAM3 (mean $\pm$ sample SD; 50 images each). DINOv3 original/DINOv3-FFA costs include encoding and both PCA reconstructions; SAM3 costs cover its automatic generation/filtering/NMS pipeline. Calibration and scoring are excluded, and stages differ. Energy is GPU board energy, not wall-plug energy; efficiency is the mean of per-image $1/J$ in $10^{-3}\,\mathrm{J}^{-1}$. Incomplete finer-focus results are omitted.}
\label{tab:cost}
\begin{tabular}{lccc}
\toprule
Method & Duration (s) & Board energy (kJ) & Efficiency ($10^{-3}\,\mathrm{J}^{-1}$) \\
\midrule
% BEGIN GENERATED TABLE cost
SAM3 automatic & 5.0 $\pm$ 0.3 & 0.32 $\pm$ 0.02 & 3.104 $\pm$ 0.181 \\
DINOv3 original & 0.317 $\pm$ 0.165 & 0.0210 $\pm$ 0.0109 & 70.072 $\pm$ 46.793 \\
DINOv3-FFA100\% & 0.328 $\pm$ 0.060 & 0.0093 $\pm$ 0.0017 & 111.835 $\pm$ 26.628 \\
DINOv3-FFA40\% & 3.1 $\pm$ 0.7 & 0.15 $\pm$ 0.05 & 7.053 $\pm$ 1.803 \\
DINOv3-FFA20\% & 11.4 $\pm$ 2.8 & 0.72 $\pm$ 0.21 & 1.480 $\pm$ 0.318 \\
DINOv3-FFA10\% & 43.8 $\pm$ 11.9 & 2.96 $\pm$ 0.83 & 0.3585 $\pm$ 0.0764 \\
DINOv3-FFA5\% & 175.8 $\pm$ 54.2 & 12.05 $\pm$ 3.72 & 0.0893 $\pm$ 0.0213 \\
DINOv3-FFA4\% & 299.5 $\pm$ 92.6 & 20.55 $\pm$ 6.36 & 0.0524 $\pm$ 0.0123 \\
DINOv3-FFA3\% & 503.5 $\pm$ 127.9 & 34.53 $\pm$ 8.76 & 0.0304 $\pm$ 0.0059 \\
DINOv3-FFA2\% & 1210.6 $\pm$ 379.5 & 83.03 $\pm$ 25.98 & 0.0130 $\pm$ 0.0031 \\
% END GENERATED TABLE cost
\bottomrule
\end{tabular}
\end{table}

\subsection{Effect of crop size on boundary preservation and energy use}
\label{sec:focus-gradient-interim}
The curves describe an implementation trade-off (Figure~\ref{fig:cost}): on average, although finer focus better preserve visible edges, it sharply increases computation. The 10\%-to-5\% transition retains a substantial gain before lower marginal energy efficiency at finer focus (Section~\ref{sec:focus-gradient-interim}). 

\begin{figure}[!htbp]
\centering\includegraphics[width=\linewidth]{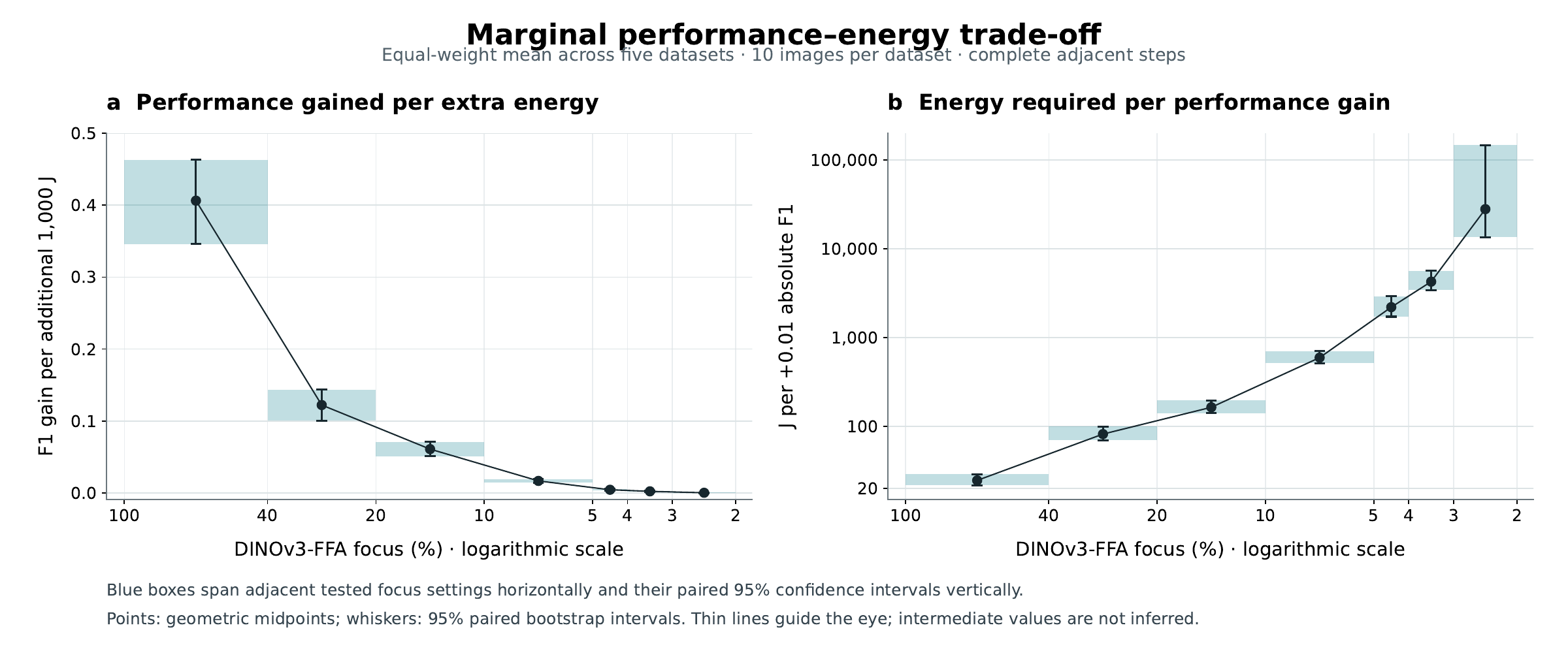}
\caption{Marginal RGB-edge F1 gain per additional 1000 J (left) and joules per $+0.01$ absolute F1 (right), calculated from paired differences with equal weight per dataset. Both horizontal axes show focus \textit{f} on a descending logarithmic scale; the left vertical axis is linear and the right is logarithmic. Each shaded rectangle spans adjacent tested focus settings horizontally and the stratified paired-image bootstrap 95\% interval vertically. Black points show observed step estimates at geometric midpoints for display; whiskers show the same intervals. Falling gain per joule and rising cost per gain describe diminishing energy efficiency, which is distinct from a zero-gain performance plateau.}
\label{fig:gradient-marginal}
\end{figure}

The retained settings represent a detail--cost choice, not a uniquely estimated biological optimum. Figure~\ref{fig:gradient-marginal} explain why finer settings were not retained as the main examples. Refining 20\% to 10\% adds mean 2242 J and 0.1368 F1, or 164 J per $+0.01$ F1. Refining 10\% to 5\% adds 9092 J and 0.1531 F1, or 594 J per $+0.01$ F1. Refining 5\% to 4\% adds another 8493 J but only 0.0386 F1. Thus 5\% retains much of the observed fine-scale edge recovery while 10\% offers a less costly view, using about one-quarter of 5\%'s extraction energy and duration. An exploratory analysis with breakpoint detection identifies 5--20\% as a suitable range considering the balance of gain and cost (Supplementary Section~\ref{app:operating-point}). 

\FloatBarrier
\section{Discussion}
This study examined whether local focus could help vision foundation models represent the dense, fine-scale visual structures found in ecological photographs. To address this question, we developed forced focused attention (FFA), implemented in \citep{vividrgb} in python, which combines representations extracted from enlarged, overlapping local crops without retraining the encoder. We evaluated the approach through visual inspection and quantitative boundary comparisons using 50 photographs across five ecological image domains. With DINOv3, FFA exposed finer appearance differences in feature displays and achieved higher mean RGB-edge boundary F1 than direct DINOv3 encoding and automatic SAM3 in every domain. Narrower focus further improved boundary preservation, although the gains diminished at finer settings while computation time and GPU energy consumption increased exponentially. These findings support local feature extraction and reconstruction as an approach to revealing fine visual cues in complex ecological imagery, with the choice of focus determining the balance between visible detail and computational cost. 

\subsection{Key findings and broader implications}
The principal finding is that changing how a frozen vision encoder views an image can substantially improve the representation of fine visual structures without retraining. FFA made individual leaves, fungal branches, and canopy boundaries more explicit in DINOv3 feature maps. A plausible explanation involves both spatial scale and contextual scope: enlarging local crops represents small structures with more encoder patches, while restricting the field of view changes the surrounding information available when their descriptors are computed. This interpretation connects with the human-vision studies introduced earlier: attention can be concentrated within a narrower spatial region \citep{eriksen1986}, global structure can take precedence over local elements in hierarchical displays \citep{navon1977}, and fine-scale information can refine an initial coarse interpretation of a scene \citep{schyns1994}. These parallels support the broader value of local inspection, while the present study demonstrates its benefit for representing fine appearance structure in ecological imagery.

FFA complements recent efforts to recover dense spatial detail from vision foundation models. \textit{FeatUp} learns an image-guided upsampler or fits an implicit representation to each image \citep{featup}, while \textit{AnyUp} learns an upsampler that transfers across encoders using crop-aligned descriptors as training targets \citep{anyup}. FFA shares tiled inference with $S^2$ \citep{s2} and local tile encoding with \textit{FeatSharp}, which learns to fuse these features with a global representation \citep{featsharp}. FFA constructs its field directly from enlarged, overlapping local encodings through inverse-aligned reflection averaging and fixed spatial weights, without learning an auxiliary upsampler or fitting an implicit reconstruction network to each image. Its contribution lies in this controllable local-focus workflow and its evaluation of appearance-boundary preservation and computational cost across five ecological domains. Beyond direct cue inspection, the reconstructed descriptors could provide image-derived teacher targets for a student trained to predict locally resolved features from a full-image input. This would offer a route to self-supervised feature distillation, related to \textit{FeatSharp}'s enhanced teacher features and \textit{LoftUp}'s local-to-global teacher--student supervision \citep{featsharp,loftup}; student training and its benefits remain to be evaluated.

Downstream ecological predictions depend on whether feature maps retain the visual variation relevant to the target question. Inadequate representation of fine visual cues could have been a technical barrier to applying general-purpose foundation models in ecological and eco-evolutionary research. FFA could help address this limitation across levels of ecological organization. At the individual level, delineating structures could support measurement of morphological traits and color variation in the source photographs, comparisons among conspecific phenotypes, genotype--phenotype analyses with matched genetic data, or screening of plant condition through yellowing and senescence \citep{schiller2021,huether2020aradeepopsis}. Hyphal density, branching, and network morphology provide related traits for studying fungal foraging strategies \citep{aguilar2022}. At the population level, instance delineation could support counts and density estimates over a known surveyed area \citep{hodgson2018drones}; at the community level, visual clusters linked to verified taxonomic labels could inform species-composition mapping \citep{schiefer2020}. At the ecosystem level, habitat classification could delineate channel and riparian units, including candidate riffle--pool structure along river continua \citep{rivasCasado2015}, and provide habitat-map inputs for landscape connectivity analyses and corridor assessment \citep{mcrae2008}. While these prospective applications require separate validation against biological or habitat reference data, the approach may unlock the utility potential of existing foundation models for various research topics.

The representation challenge illustrated by ecological photographs also arises in other domains where many locally distinct structures coexist within a single field of view. In medical imaging, histological sections contain diverse cell types, tissue substructures, and pathological features that can be distinguished using self-supervised representations \citep{cisternino2024}; color fundus photographs combine small retinal lesions, including microaneurysms, hemorrhages, and exudates, with normal anatomical structures \citep{porwal2018idrid}. Beyond biology, metallographic micrographs contain particles, grain-boundary carbide networks, and matrix regions whose delineation enables quantitative characterization of material microstructure \citep{decost2019}. Crowded aerial images similarly place small vehicles and ships alongside larger infrastructure, with substantial variation in scale and orientation \citep{xia2018dota}. Across these settings, structures occupying little image area may still carry information relevant to the target question. These parallels motivate testing whether controlled local encoding can make such distinctions more accessible in the descriptors of a frozen encoder. FFA's transfer to these domains remains to be evaluated, including how locally resolved features can be combined with the broader spatial context needed for interpretation.

\subsection{Limitations}
While seeing the potential benefits of FFA application, several choices in the study design limit the conclusions that can be drawn robustly. The primary quantitative endpoint measures agreement between boundaries of clustered descriptors and RGB-derived brightness edges, which may reflect texture, shadows, or artifacts and can miss biologically meaningful distinctions. Such a proxy requires validation against the intended biological task \citep{reinke2024metrics}. Scores also depend on the retained PCA dimensions, cluster number, and edge-matching rules; the sensitivity analyses do not substitute for evaluating species identification, trait measurement, counting, or other downstream tasks. The 50 selected photographs provide examples across five domains, but do not establish generalization across representative datasets or acquisition conditions. Transfer difficulties documented in camera-trap recognition illustrate why independent environments require explicit testing \citep{beery2018terra}. Evidence centers on DINOv3, with only a five-image SAM3 encoder extension; the retained appearance-boundary results do not establish biological segmentation accuracy. Matched comparisons with feature-reconstruction methods such as \textit{FeatUp} and \textit{AnyUp} \citep{featup,anyup} are also missing. 

FFA also has limitations inherent in its construction. It is a simple inference procedure with manually specified crop geometry and fixed aggregation weights, and neither the frozen encoder nor the reconstruction is optimized for a downstream objective. Overlapping crops and reflection averaging require many forward passes: narrower focus increased computing time and GPU energy while yielding diminishing gains in boundary F1. Avoiding retraining therefore does not imply efficient inference; benefits need to be assessed alongside computational cost \citep{schwartz2020greenai}. System-level energy accounting includes components beyond the GPU \citep{henderson2020energy}; our measurements exclude CPU and host-memory energy and other workflow stages. Restricting each view also withholds the surrounding image context, potentially weakening information about whole-object identity and larger spatial relationships. In the over-tiling experiment of \textit{FeatSharp}, finer tiling produced greater discrepancies from whole-image features, which the authors associated with insufficient context. Its learned global--local fusion illustrates a way to integrate local detail and broader organization \citep{featsharp}. 

\subsection{Future work}
Generally, relevant information may span a continuum of spatial scales, with their relative importance varying across structures, image locations, and prediction targets. A fixed focus fraction does not represent a consistent physical scale across photographs or identify the task-dependent contributions of different spatial scales. A useful extension would sample a range of spatial scopes and learn their relative contributions, following scale-aware weighting approaches \citep{chen2016scale}, while testing whether adaptive sampling can reduce computation without sacrificing useful context. 

A factorial comparison is needed to separate effective resolution, contextual scope, and aggregation under comparable computational budgets. Independent benchmarks with task-specific biological annotations and evaluation settings fixed in advance are needed to establish the wider utility of FFA.

\section{Conclusion}
Forced focused attention retrieves dense visual cues by changing the scale and contextual scope of a frozen encoder's inputs, then reconstructing their descriptors into a common image field. With DINOv3, enlarged local crops increase mean RGB-edge boundary F1 over both declared comparators in all five selected domains (Table~\ref{tab:f1}).  The study offers an inspectable, reproducible workflow, released as the alpha Python library \texttt{vividRGB} 0.1.0. 

\section*{Data and code availability}
The Python library \texttt{vividRGB} version 0.1.0 is an alpha release available on \href{https://pypi.org/project/vividRGB/0.1.0/}{PyPI} and at \href{https://github.com/masahiroryo-1988/vividRGB/releases/tag/v0.1.0}{the versioned GitHub release}. Install with \texttt{pip install vividRGB}; the import is \texttt{vividRGB} and the command-line executable is \texttt{vividrgb}. The optional DINO and SAM3 adapters require their documented dependencies and upstream checkpoint access. The release implements local reconstruction, PCA/varimax displays, fixed or automatic K-means, penalized GMM, boundary evaluation, and a custom dense-encoder interface.  \citep{vividrgb}.

\section*{Acknowledgments and declarations}
This study was supported by the Bundesministerium für Bildung und Forschung (BMBF) project “KI und Citizen Science gestütztes Monitoring von zertifizierten Biodiversitätsprojekten (KICS-Zert)” (16LW0441). I acknowledge MOIN/KICS-Zert photograph contributors and the creators of the fungal-network, NEON, Coralscapes, and PMID2019 resources, as well as the DINOv3, SAM3, and scientific Python developers. There is no conflict of interest.

\FloatBarrier

\clearpage
\appendix
\setcounter{table}{0}
\setcounter{figure}{0}
\setcounter{equation}{0}
\renewcommand{\thesection}{S\arabic{section}}
\renewcommand{\thetable}{S\arabic{table}}
\renewcommand{\thefigure}{S\arabic{figure}}
\renewcommand{\theequation}{S\arabic{equation}}
\renewcommand{\theHtable}{supp.\arabic{table}}
\renewcommand{\theHfigure}{supp.\arabic{figure}}
\renewcommand{\theHequation}{supp.\arabic{equation}}
\section*{Supplementary material}
\addcontentsline{toc}{section}{Supplementary material}
\section{Details in technical implementation}\label{app:evaluation-details}\label{app:boundary-details}\label{app:reproducibility}

\subsection{Frozen encoders and automatic-mask comparator}\label{app:encoder-settings}
\paragraph{DINOv3.}
All DINOv3 conditions use the frozen \nolinkurl{facebook/dinov3-vitb16-pretrain-lvd1689m} checkpoint, revision \nolinkurl{5931719e67bbdb9737e363e781fb0c67687896bc}, without training, fine-tuning, or target annotations. Patch stride is 16 pixels and each last-hidden-state patch descriptor has 768 components; class and register tokens are discarded. ``DINOv3 original'' means one orientation at the entire working image's native resolution, not the checkpoint processor's default thumbnail or the dimensions of an uncropped source photograph. Image-wide context is available to this control. Patch stride specifies its token lattice, not the contextual extent of each output descriptor.

Input RGB is divided by 255 and standardized with means $(0.485,0.456,0.406)$ and standard deviations $(0.229,0.224,0.225)$. The processor's default $224\times224$ resize is not applied. No additional L2 normalization is applied to output descriptors. Inference uses evaluation mode, no gradients, scaled-dot-product attention, and FP16 autocasting. Matrix-multiplication TF32 is disabled; cuDNN TF32 retains the framework default. Reflection averages accumulate in float32 and are cast to FP16 before projection; PCA projection and reconstruction use float32.

\paragraph{Automatic SAM3 masks.}
The automatic comparator uses the Transformers \nolinkurl{mask-generation} pipeline with \texttt{facebook/sam3}, revision \nolinkurl{3c879f39826c281e95690f02c7821c4de09afae7}, through \nolinkurl{Sam3TrackerModel} \citep{sam3,hfsam3}. Only the image is supplied by the user; the pipeline internally generates a $32\times32$ grid of point prompts. ``Automatic'' therefore does not mean that the architecture never uses prompts. No human-selected text, point, box, mask, object name, or ecological category is provided. Settings are four points per batch, no additional crop layers, predicted-IoU threshold 0.88, stability threshold 0.95, logit threshold 0, stability offset 1, and mask-box NMS threshold 0.7. The processor resizes to $1008\times1008$, divides RGB by 255, and uses channel mean and standard deviation 0.5. Inference is FP32 with matrix-multiplication TF32 disabled. Highest predicted mask IoU wins overlaps; uncovered pixels remain label 0. Custom-grid and concept-prompted results are excluded. These automatic masks are distinct from the raw SAM3 image-encoder partitions in Section~\ref{app:sam3-ffa}.

\subsection{Crop geometry, reflections, and reconstruction}\label{app:reconstruction-details}
For a working image of width $W$, height $H$, and short side $L=\min(W,H)$, focus fraction $f$, minimum side $c_{\min}$, and overlap $\alpha$ define
\begin{equation}
 c_f=\max\{c_{\min},\operatorname{round}(fL)\},\qquad
 s_f=\max\{1,\operatorname{round}[(1-\alpha)c_f]\}.
 \label{eq:geometry}
\end{equation}
Each axis uses regular starts plus a final edge-aligned start, deduplicated and sorted for complete coverage. Realized interior overlap is $1-s_f/c_f$; rounding and final windows can change it. Focus is a fraction of short-side \emph{length}, not area: a 5\% crop of a square image occupies approximately 0.25\% of its area.

The five-domain comparison uses $f=0.05$ or $0.10$  (for the primary setting), $c_{\min}=16$, nominal $\alpha=0.75$, and bicubic rescaling to $384\times384$ before RGB standardization. With patch stride 16, this gives a $24\times24$ token grid. For a $2048\times2048$ image, crop sides are 102 and 205 pixels and nominal source-coordinate token pitches are 4.25 and 8.54 pixels, respectively, versus the control's 16-pixel patch stride. Interpolation adds no observed image detail. The crop excludes outside RGB during that inference pass, but reconstruction combines overlapping contexts; a final descriptor does not have one fixed crop-sized neighborhood. No attention-weight analysis establishes a causal account of internal model behavior. The 75\% overlap was chosen after inspection, not established as a universal or quantitatively optimal value.

For crop $I_b$, frozen encoder $F$, and transformations $\mathcal G$ comprising identity, horizontal reflection, vertical reflection, and both, inverse-align the token grids before averaging:
\begin{equation}
 \bar Z_b=\frac14\sum_{g\in\mathcal G}T_g^{-1}[F(T_g I_b)].\label{eq:flip}
\end{equation}
This inverse-aligned averaging follows the approach of Docherty et al.\ \citep{docherty2026}, applied here to each rescaled crop. A shared PCA mean $\mu$ and basis $B\in\R^{768\times d}$ give $Y_b=(\bar Z_b-\mu)B$. Bilinear interpolation with \texttt{align\_corners=False} places these coordinates on the crop's source grid. For a length-$c$ NumPy Hann window $h_c$, the taper is
\begin{equation}
 w_b(u,v)=\max\{h_c(u),0.05\}\max\{h_c(v),0.05\};
\end{equation}
its per-axis floor is 0.05 and its two-dimensional minimum is 0.0025. The reconstructed field is
\begin{equation}
 Y_f(x)=\frac{\sum_{b:x\in b}w_b(x)\,U_b[Y_b](x)}{\sum_{b:x\in b}w_b(x)},\label{eq:blend}
\end{equation}
where $U_b$ denotes bilinear placement and the denominator is checked positive everywhere. Features, not colors or cluster IDs, are averaged. For a fixed affine PCA projection and normalized weights, projection and averaging commute apart from floating-point differences; the result still occupies a truncated subspace. Reflection averaging addresses orientation sensitivity, while overlap and tapering reduce abrupt changes in supplied context.

\subsection{Fixed-count partitions}\label{app:fixed-partitions}
Primary DINO-derived scores use independent K-means models in all 16 PCA coordinates, with $K=6$, ten initializations, and seed 42. A NumPy default generator with seed 45 uniformly samples 8192 distinct full-image pixel positions without replacement for each image, reused across treatments. These positions select features; coordinates are not clustering variables. This fitting sample is separate from PCA-token sampling and the evaluation mask. The objective is
\begin{equation}
 \min_{a_i,m_k}\sum_i\|Y(x_i)-m_{a_i}\|_2^2.
\end{equation}
Centers are not shared, and every source-grid position is assigned to its nearest center. K-means imposes no equal-area constraint. $K=12$ is a recorded sensitivity. Fixed-$K$ F1 scores are not scores of the later automatic-$K$ maps. A predicted boundary pixel differs in label from its right or lower neighbor.

\subsection{Boundary reference, scoring, and appearance controls}\label{sec:boundary-methods}
The primary endpoint evaluates an encoder--partition combination against strong RGB appearance transitions, without requiring arbitrary cluster IDs to correspond. On stored gamma-encoded RGB channels scaled to $[0,1]$, grayscale is
\begin{equation}
 I_{\rm gray}=0.2126r_{\rm RGB}+0.7152g_{\rm RGB}+0.0722b_{\rm RGB}.
\end{equation}
These Rec.\ 709-style weights \citep{rec709} are applied without linearization, giving a luma-like proxy, not physical luminance. At full working resolution, Gaussian smoothing ($\sigma=1$ pixel) precedes horizontal and vertical Sobel derivatives divided by 8. The magnitude $G=(G_x^2+G_y^2)^{1/2}$ defines reference edges at or above $\max(10^{-6},q_{0.85})$, where $q_{0.85}$ is that image's eligible-gradient 85th percentile. Thus approximately the strongest 15\% of gradients are selected, with ties potentially including more. No thinning or nonmaximum suppression is applied; reference edges may be bands rather than single-pixel contours.

For five-domain RGB scoring, the valid domain $V$ is the entire image eroded by two pixels. Restrict predicted boundaries $P$ and reference $E$ to $V$. With $D_t$ denoting Manhattan dilation (SciPy default cross connectivity) of radius $t=2$ working pixels,
\begin{equation}
 P_t=\frac{|P\cap D_t(E)|}{|P|},\qquad
 R_t=\frac{|E\cap D_t(P)|}{|E|},\qquad
 F_{1,t}=\frac{2P_tR_t}{P_t+R_t}.\label{eq:f1}
\end{equation}
Empty denominators yield zero. This is tolerant set alignment, not one-to-one matching or region IoU. Multiple predicted pixels can cover the same reference edge. Boundary density $|P|/|V|$ is recorded because dense or fragmented partitions can raise recall; precision does not remove all density dependence.

The RGB-only positive control applies independent $K=6$ K-means to the three RGB/255 channels, with the same 8192 locations, ten initializations, seed 42, full-resolution prediction, masks, and tolerance. There is no encoder, PCA, channel standardization, spatial feature, or display transformation. Its label boundaries are compared with the unclustered-image gradient reference, not with themselves. Shared RGB appearance makes this an informative positive control, not an independent biological reference or a theoretical F1 upper bound.

\subsection{Continuous PCA and varimax displays}\label{sec:display-methods}
Ordinary PCA displays map the first three scores to RGB using shared calibration-token 1st--99th percentile limits. Varimax instead rotates the projection-weight basis $B$, not covariance loadings $B\operatorname{diag}(\sqrt\eta)$, separately at $d=3$ or $16$ \citep{kaiser1958}. For optimization only, Kaiser row normalization divides row $i$ by $\max(\|B_{i\cdot}\|_2,10^{-15})$, giving $A$. Orthogonal $R$ maximizes:
\begin{equation}
 V(R)=\sum_{j=1}^{d}\left[\sum_{i=1}^{768}(AR)_{ij}^{4}
 -\frac1{768}\left\{\sum_{i=1}^{768}(AR)_{ij}^{2}\right\}^{2}\right],\quad R^\top R=I.
\end{equation}
DINOv3 varimax displays use the first three rotated axes and shared 2nd--98th percentile limits. 

For the additional vegetation illustration in Figures~2 and~3, the full photograph used in Figure~1A was oriented and downsampled to a $961\times1280$ working RGB image without cropping. For display only, a common square crop (all 961 columns and zero-based rows 100--1060, inclusive) was taken from the RGB image and every derived map, making the first-row panels as wide as those in the second row; encoder inference, PCA/varimax, GMM fitting, and SAM3 mask generation used the uncropped working image. Original DINOv3, FFA5\%, and FFA10\% fields were projected into one image-specific centered, unwhitened PCA16 basis calibrated with 2048 tokens from each of six treatments (native, whole-image reflections, and 5\%, 10\%, 20\%, and 40\% focus). The vegetation display uses an orthogonal rotation of that full 16-dimensional basis and shared 2nd--98th percentile limits from equal samples of the three displayed treatments; the four original benchmark rows retain their archived limits, which also pooled FFA20\%. Separate full-covariance GMMs used the same $K=1,\ldots,12$ candidate range and $\lambda=4$ penalty as the other rows, while SAM3 used the same automatic-mask settings. This photograph is an illustration and is not among the 50 scored images.

\subsection{Configuration and reproducibility}\label{app:configuration}
Table~\ref{tab:parameters} identifies the controls used in this study. 

\begin{table}[htbp]
\centering\small
\caption{Essential controls by experiment. Encoder architecture is fixed by the checkpoint; crop geometry, calibration, partitioning, and evaluation are experimental choices. Selected methodological details are given in the surrounding supplementary sections.}\label{tab:parameters}
\begin{tabular}{@{}>{\raggedright\arraybackslash}p{34mm}>{\raggedright\arraybackslash}p{117mm}@{}}
\toprule
Experiment or control & Recorded settings \\
\midrule
Primary DINOv3 comparison & Native single-orientation control; FFA 5\% and 10\%; 384-pixel input, 75\% overlap, four reflections; six-treatment shared PCA16; independent $K=6$ (12 sensitivity). \\
Reconstruction & Crop minimum 16; bilinear placement; per-axis Hann floor 0.05; positive coverage normalization; no token L2 normalization or feature whitening. \\
Automatic SAM3 & 1008-pixel input; internal $32\times32$ point grid; batch 4; no crop layers; IoU 0.88, stability 0.95, NMS 0.7; highest-IoU overlap priority. \\
RGB-edge endpoint & Gamma-encoded grayscale; Gaussian $\sigma=1$; Sobel/8; threshold $\max(10^{-6},q_{0.85})$; two-pixel erosion and Manhattan tolerance. \\
Automatic grouping & Long-side-256 grids; PCA $d=3,16$; K-means $K=3$--12; full-covariance GMM $K=1$--12; post hoc $\lambda=4$; sensitivity $\lambda=1$--64. \\
Crop-size analysis & Separate calibration; eight complete 50-image cohorts, 100\%--2\%; four reflections; two reconstruction branches in measured DINOv3 cost. \\
\end{tabular}
\end{table}
\FloatBarrier

\section{Comparative results and controls}
\subsection{Paired comparisons and their inferential scope}\label{app:contrasts}
The primary comparisons use ten image-level differences per domain, not pixels or crops as replicates. Five domains, two FFA scales, and two baselines give 20 contrasts. Holm correction covers all 20 tests \citep{holm1979}. Descriptive paired-percentile intervals use 20,000 image-bootstrap draws (seed 20261005), without multiplicity adjustment. Table~\ref{tab:contrasts} retains all comparisons. 

% Complete source values are inserted below without recalculation.
\begin{table}[htbp]
\centering\small\setlength{\tabcolsep}{3pt}
\caption{Complete paired differences in RGB-edge boundary F1: FFA minus baseline, $n=10$ per domain. DINOv3 denotes the native original control; SAM3 denotes automatic masks. Wins are strictly positive differences. Intervals are descriptive and unadjusted. Values are retained from the frozen per-image results.}\label{tab:contrasts}
\begin{tabular}{@{}lllr l r rr@{}}
\toprule
Domain & Focus & Baseline & $\Delta$ & 95\% interval & Wins & Raw $p$ & Holm $p$ \\
\midrule
Grassland & 5\% & DINOv3 & 0.535 & [0.460, 0.604] & 10/10 & 0.00195 & 0.03906 \\
Grassland & 5\% & SAM3 & 0.575 & [0.505, 0.642] & 10/10 & 0.00195 & 0.03906 \\
Grassland & 10\% & DINOv3 & 0.380 & [0.326, 0.434] & 10/10 & 0.00195 & 0.03906 \\
Grassland & 10\% & SAM3 & 0.421 & [0.367, 0.476] & 10/10 & 0.00195 & 0.03906 \\
Fungi & 5\% & DINOv3 & 0.214 & [0.144, 0.289] & 10/10 & 0.00195 & 0.03906 \\
Fungi & 5\% & SAM3 & 0.257 & [0.181, 0.327] & 10/10 & 0.00195 & 0.03906 \\
Fungi & 10\% & DINOv3 & 0.167 & [0.111, 0.226] & 10/10 & 0.00195 & 0.03906 \\
Fungi & 10\% & SAM3 & 0.210 & [0.143, 0.274] & 10/10 & 0.00195 & 0.03906 \\
Forest & 5\% & DINOv3 & 0.516 & [0.433, 0.587] & 10/10 & 0.00195 & 0.03906 \\
Forest & 5\% & SAM3 & 0.509 & [0.417, 0.588] & 10/10 & 0.00195 & 0.03906 \\
Forest & 10\% & DINOv3 & 0.326 & [0.247, 0.403] & 10/10 & 0.00195 & 0.03906 \\
Forest & 10\% & SAM3 & 0.319 & [0.237, 0.400] & 10/10 & 0.00195 & 0.03906 \\
Coral & 5\% & DINOv3 & 0.395 & [0.310, 0.479] & 10/10 & 0.00195 & 0.03906 \\
Coral & 5\% & SAM3 & 0.400 & [0.309, 0.490] & 10/10 & 0.00195 & 0.03906 \\
Coral & 10\% & DINOv3 & 0.225 & [0.181, 0.275] & 10/10 & 0.00195 & 0.03906 \\
Coral & 10\% & SAM3 & 0.230 & [0.181, 0.285] & 10/10 & 0.00195 & 0.03906 \\
Phytoplankton & 5\% & DINOv3 & 0.427 & [0.353, 0.490] & 10/10 & 0.00195 & 0.03906 \\
Phytoplankton & 5\% & SAM3 & 0.365 & [0.292, 0.431] & 10/10 & 0.00195 & 0.03906 \\
Phytoplankton & 10\% & DINOv3 & 0.260 & [0.176, 0.339] & 10/10 & 0.00195 & 0.03906 \\
Phytoplankton & 10\% & SAM3 & 0.198 & [0.114, 0.278] & 9/10 & 0.00391 & 0.03906 \\
\bottomrule
\end{tabular}
\end{table}
\FloatBarrier

\section{Sensitivity and computational cost}\label{app:cluster-sensitivity}
\subsection{Automatic clustering}\label{sec:group-methods}
Later clustering samples dense PCA fields at cell centers on aspect-preserving grids with longest side 256. These are interpolated descriptors, not independently encoded pixels. Algorithms use identical grids with $d=3$ or $16$, without added RGB, XY coordinates, whitening, or unit-vector normalization.

Automatic K-means considers $K=3,\ldots,12$. An $8\times8$ spatial block layout is shuffled into three folds; each fits at most 3072 locations and evaluates silhouette on at most 1024 held-out locations \citep{rousseeuw1987}. Among candidates within one standard error of the highest mean silhouette, selection favors $K$ closest to five, then smaller $K$. This standard error is a three-fold model-selection tolerance, not calibrated independent-sample uncertainty. Final fits use at most 8192 fixed locations and ten initializations. The method contains an explicit preference, not preference-free discovery.

Full-covariance GMMs represent appearance density as $p(y)=\sum_{k=1}^{K}\pi_k\mathcal N(y;m_k,\Sigma_k)$, with nonnegative weights summing to one; maximum posterior responsibility gives a partition. Candidates $K=1,\ldots,12$ use the same fit locations, float64, three initializations, seed 42, diagonal ridge $10^{-5}$, tolerance $10^{-3}$, and 300 iterations with warm continuation to 600 if needed. Only converged fits are eligible. With summed fitted log density $\log L_K$, selection minimizes
\begin{align}
 S_\lambda(K)&=-2\log L_K+\lambda p_K\log n,\label{eq:penalty}\\
 p_K&=K\left[d+\frac{d(d+1)}2\right]+K-1.
\end{align}
At $d=16$, $p_K=153K-1$. The sweep is $\lambda=1,2,4,8,16,32,64$; ties favor smaller $K$. $\lambda=1$ is ordinary BIC \citep{schwarz1978}; larger values strengthen the BIC-style penalty, not a Bayesian concentration prior. Spatial dependence and finite mixtures limit regular independent-observation interpretations. Criterion values are not compared across dimensions. 

Increasing $\lambda$ reduces counts (Table~\ref{tab:penalty}) but can collapse useful distinctions. At $\lambda=4$, mean counts are 5.34 and 6.34, with no cap selections. Automatic K-means instead selects means 3.60 and 3.58 (both medians 3), with ranges 3--6 and 3--5; its preference toward five does not force five, and its lower bound can be active. Note that this experiment was done additionally to the main study topic and not relevant directly to the F1 score assessment.

\begin{table}[htbp]
\centering\small
\caption{Effect of the strengthened BIC-style penalty at 16 dimensions. Frequencies are empirical percentages over the same 50 images; ``cap'' is selection of the maximum candidate $K=12$. The retained exploratory setting is $\lambda=4$.}
\label{tab:penalty}
\begin{tabular}{r rrrr rrrr}
\toprule
& \multicolumn{4}{c}{DINOv3-FFA5\%} & \multicolumn{4}{c}{DINOv3-FFA10\%} \\
\cmidrule(lr){2-5}\cmidrule(lr){6-9}
$\lambda$ & Mean $K$ & Median & $K=4$--6 & Cap & Mean $K$ & Median & $K=4$--6 & Cap \\
\midrule
% BEGIN GENERATED TABLE penalty
1 & 11.14 & 11 & 0\% & 48\% & 11.90 & 12 & 0\% & 96\% \\
2 & 8.10 & 8 & 14\% & 8\% & 10.26 & 11 & 4\% & 36\% \\
4 & 5.34 & 5 & 74\% & 0\% & 6.34 & 6 & 64\% & 0\% \\
8 & 3.38 & 3 & 34\% & 0\% & 4.04 & 4 & 56\% & 0\% \\
16 & 2.38 & 2 & 8\% & 0\% & 2.64 & 3 & 10\% & 0\% \\
32 & 1.50 & 1 & 0\% & 0\% & 1.52 & 1 & 0\% & 0\% \\
64 & 1.26 & 1 & 0\% & 0\% & 1.12 & 1 & 0\% & 0\% \\
% END GENERATED TABLE penalty
\bottomrule
\end{tabular}
\end{table}

The $\lambda=4$ response is not invariant to retained dimension (Table~\ref{tab:dimensions-condensed}). Three dimensions discard information and alter the parameter penalty. Restricting clustering to display axes may ease visual inspection but is not established as more accurate; rotating all sixteen axes and displaying three is different from fitting only three dimensions.

\begin{table}[htbp]
\centering\small
\caption{Full-covariance GMM selection at $\lambda=4$ with 3 versus 16 retained PCA dimensions. Each row contains 50 images. Criterion values are not compared across dimensions.}\label{tab:dimensions-condensed}
\begin{tabular}{@{}rlrrrr@{}}
\toprule
$d$ & Features & Mean $K$ & Median & $K=4$--6 & $K=12$ \\
\midrule
3 & DINOv3 original & 10.02 & 11 & 10\% & 42\% \\
3 & DINOv3-FFA5\% & 5.60 & 5.5 & 60\% & 0\% \\
3 & DINOv3-FFA10\% & 6.58 & 6 & 46\% & 4\% \\
16 & DINOv3 original & 11.90 & 12 & 0\% & 92\% \\
16 & DINOv3-FFA5\% & 5.34 & 5 & 74\% & 0\% \\
16 & DINOv3-FFA10\% & 6.34 & 6 & 64\% & 0\% \\
\bottomrule
\end{tabular}
\end{table}

\subsection{Measured duration, board energy, and marginal gain}\label{app:focus-cost-details}
Measurements use one NVIDIA RTX 4000 SFF Ada Generation GPU (approximately 20 GiB). Model loading and warm-up are excluded. Local extraction includes crop preparation, four encoder orientations, projection, interpolation, and blending. Calibration and clustering/scoring/rendering are outside this stage. In the original comparison, DINOv3 original duration combines recorded encoder time with projection/interpolation; its encoder output was reused from cache and complete-stage energy was unavailable. The crop-size analysis instead directly measures original-control energy and includes \emph{two} PCA projection/reconstruction branches for DINOv3: the crop-size and primary-comparison calibrations. Checkpoint I/O and power-bracketing waits are excluded.

Automatic SAM3 cost covers its separately measured generation/filtering/NMS pipeline, excluding partition assembly, scoring, and rendering. The main-paper cost comparison uses newly measured DINOv3 costs and reused automatic SAM3 costs. These differing stage boundaries are operational implementation costs, not total-workflow or compute-normalized architecture comparisons.

Board power from \texttt{nvidia-smi power.draw} is sampled approximately every 200 ms and integrated by
\begin{equation}
 J\approx\sum_j\frac{P(t_j)+P(t_{j+1})}{2}(t_{j+1}-t_j),\qquad \epsilon_J=1/J.
\end{equation}
No idle baseline is subtracted. CPU, host memory, wall-plug losses, and carbon accounting are excluded. Efficiency summaries average per-image reciprocals, not the reciprocal of mean energy. Timings are single runs rather than replicated hardware benchmarks; short control stages are especially sensitive to sample spacing and transient board state.

For adjacent coarser-to-finer settings, paired changes are averaged within each domain and then equally across the five domains. Write these means as $\Delta E$ and $\Delta F$. Marginal gain and cost are
\begin{equation}
 G_{1000}=1000\frac{\Delta F}{\Delta E},\qquad
 C_{0.01}=0.01\frac{\Delta E}{\Delta F}.\label{eq:supp-marginal-energy}
\end{equation}
The first requires positive additional energy; the second additionally requires positive gain and scales an observed step, not an assertion that another 0.01 gain is attainable. Ratios use paired means rather than unstable individual ratios. Within-domain paired-image bootstraps use 20,000 draws (seed 20261007) for 95\% intervals; ratio intervals are withheld when their denominator interval reaches zero.

\subsection{Crop-size choice with a breakpoint analysis}\label{app:operating-point}
These post hoc diagnostics use the same complete 400 observations, with equal weight for domain means, unchanged calibration and $K=6$ scoring. They give descriptive context for the selected settings, not an independently validated tuning rule. For each declared horizontal coordinate $x$, equal-weight ordinary least squares fits
\begin{equation}
 \overline{F1}(x)=\beta_0+\beta_1x+\beta_2(x-\tau)_+,\qquad (a)_+=\max(0,a).\label{eq:operating-hinge}
\end{equation}
The observed eligible knot with least residual sum of squares is selected; no interpolated optimum is sought. Each segment includes at least three settings, counting the knot in both. Full-range candidates are 20\%, 10\%, 5\%, and 4\%; among the six settings from 20\% through 2\%, only 5\% and 4\% qualify. Coordinates are $\log_{10}$ mean energy, $\log_{10}(100/f_{\%})$, and linear mean energy; linear rescaling leaves fit and knot unchanged. Negative slope change denotes deceleration, positive change acceleration. Five thousand stratified paired-image bootstraps (seed 20261009) resample ten identifiers within each domain, moving each image's scores and energies at all eight settings together. Means, energy coordinates, and knots are recomputed. Frequencies and percentile intervals are conditional on the selected cohort and candidate settings.

\begin{table}[htbp]
\centering\small\setlength{\tabcolsep}{4pt}
\caption{Hinge sensitivity to coordinate and included focus range. Frequency is bootstrap reselection of the observed knot, not the probability of a biological optimum.}\label{tab:operating-hinge}
\begin{tabular}{@{}llrll@{}}
\toprule
Range & Coordinate & Knot & Slope change & Frequency \\
\midrule
100\%--2\% & $\log_{10}$ mean energy & 20\% & Acceleration & 100.0\% \\
100\%--2\% & $\log_{10}$ inverse focus & 20\% & Acceleration & 100.0\% \\
100\%--2\% & Linear mean energy & 5\% & Deceleration & 62.3\% \\
40\%--2\% & $\log_{10}$ mean energy & 4\% & Deceleration & 99.3\% \\
20\%--2\% & $\log_{10}$ mean energy & 4\% & Deceleration & 93.0\% \\
\bottomrule
\end{tabular}
\end{table}

Within the finer range, the 4\% hinge reduces fitted slope from 0.233 to 0.073 absolute F1 per $\log_{10}$ joule; reselected-knot bootstrap intervals are $[0.214,0.256]$ and $[0.042,0.113]$. The full-range linear-energy alternative is 10\% in 37.7\% of resamples. A separate maximum-deviation-above-endpoint-chord diagnostic after normalizing axes to $[0,1]$ selects 5\% for linear energy, 3\% for log energy, and 4\% for log inverse focus. There is therefore no coordinate-independent knee.

All eight observed mean conditions are nondominated in mean energy versus mean F1. Energy-penalized utility compares observed mean F1 minus an application-specified penalty on mean energy; preference intervals are intersections of those measured utility lines, not hinge locations. Lower energy penalties favor finer settings. No unique optimum exists without an energy preference. These results concern one GPU, an appearance endpoint, and stages excluding calibration and clustering.
\FloatBarrier

\section{Exploratory SAM3 image-encoder tests}\label{app:sam3-ffa}
\subsection{Scope and matched protocol}\label{sec:sam3-methods}
This extension tests application of FFA to a second frozen encoder, not two complete segmentation systems. It uses five existing illustrations: \texttt{moin\_01}, \texttt{fungal\_network\_01}, \texttt{neon\_01}, \texttt{coralscapes\_01}, and \texttt{pmid\_01}. They were retained from the stored figure selection before their new SAM3 scores; they are not a random sample or replicated domain validation. All five-case summaries are descriptive, without population confidence intervals or significance tests.

The same pinned \texttt{facebook/sam3} checkpoint supplies \nolinkurl{vision_encoder.backbone.last_hidden_state}, before prompt fusion, neck processing, and mask decoding. A $1008\times1008$ input yields $72\times72$ descriptors, patch side 14, and 1024 channels. No text, exemplar, point, box, or mask is supplied. This raw image-only experiment is separate from the automatic-mask pipeline, which generates internal point prompts. SAM3 original uses one orientation and official square resizing, RGB/255 rescaling, and channel mean/SD 0.5. FFA10\% and FFA5\% use source-coordinate short-side crops, minimum 16, bicubic resizing to 1008, four inverse-aligned reflections, nominal 75\% overlap, and the same floored Hann reconstruction. Crop sides at 10\%/5\% are 205/102 for the three $2048\times2048$ images, 136/68 for forest, and 102/51 for coral. Effective fractions span 9.9609--10.0098\% and 4.9805--5.0000\%; none reaches the minimum. 

\subsection{Appearance-boundary outcomes and density}
The RGB reference and two-pixel tolerance are unchanged. Matched DINOv3 entries use the crop-size calibration on these five images, not the primary fifty-image comparison's calibration. Table~\ref{tab:sam3-ffa-cases} shows all five cases; Table~\ref{tab:sam3-ffa-diagnostics} separates edge coverage, precision, density, and the three-axis sensitivity.

\begin{table}[htbp]
\centering\small\setlength{\tabcolsep}{4pt}
\caption{Primary RGB-edge boundary F1 for the five matched illustrations ($K=6$, 16 PCA coordinates). Each domain entry is one image; the final row is its descriptive five-example mean. SAM3 columns are image-encoder partitions, not automatic masks.}
\label{tab:sam3-ffa-cases}
\begin{tabular}{lrrrrrr}\toprule
& \multicolumn{3}{c}{DINOv3} & \multicolumn{3}{c}{SAM3 image encoder} \\
Domain & Original & FFA10\% & FFA5\% & Original & FFA10\% & FFA5\% \\
\midrule
Grassland & 0.0915 & 0.4837 & 0.5787 & 0.2487 & 0.4733 & 0.4799 \\
Fungal networks & 0.0314 & 0.2172 & 0.3346 & 0.1258 & 0.3040 & 0.3439 \\
Airborne forest & 0.0317 & 0.3979 & 0.6096 & 0.2997 & 0.5789 & 0.6062 \\
Coral reefs & 0.0511 & 0.2172 & 0.4144 & 0.2512 & 0.4522 & 0.4923 \\
Phytoplankton & 0.0240 & 0.0654 & 0.1637 & 0.1316 & 0.4004 & 0.4496 \\
\midrule Five-example mean & 0.0459 & 0.2763 & 0.4202 & 0.2114 & 0.4417 & 0.4744 \\
\bottomrule\end{tabular}
\end{table}

\begin{table}[htbp]
\centering\small\setlength{\tabcolsep}{4pt}
\caption{Descriptive means across the same five cases. Precision, recall, F1, and boundary density use the primary 16-coordinate partition. The last column uses a separately fitted $K=6$ partition of the first three varimax axes. Mean F1 is the mean of image-level F1, not the harmonic mean of the displayed mean precision and recall. Density is the fraction of eligible image pixels marked as predicted boundaries.}
\label{tab:sam3-ffa-diagnostics}
\begin{tabular}{lrrrrr}\toprule
Method & Precision & Recall & F1 & Density & 3-axis F1 \\
\midrule
DINOv3 original & 0.4853 & 0.0244 & 0.0459 & 0.0040 & 0.1792 \\
DINOv3-FFA10\% & 0.4368 & 0.2184 & 0.2763 & 0.0420 & 0.2864 \\
DINOv3-FFA5\% & 0.4764 & 0.3984 & 0.4202 & 0.0761 & 0.3843 \\
SAM3 image encoder & 0.3403 & 0.1559 & 0.2114 & 0.0445 & 0.2654 \\
SAM3-FFA10\% & 0.3083 & 0.8012 & 0.4417 & 0.4064 & 0.5317 \\
SAM3-FFA5\% & 0.3227 & 0.9230 & 0.4744 & 0.5895 & 0.5284 \\
\bottomrule\end{tabular}
\end{table}

SAM3-FFA improves primary F1 relative to its own original-image control in all five cases: means increase from 0.2114 to 0.4417 at 10\% and 0.4744 at 5\% (gains 0.2303 and 0.2630). Five-percent focus exceeds 10\% in each primary case, but adds only 0.0326 on average. Matched DINOv3 means are 0.0459, 0.2763, and 0.4202. Cross-backbone values compare workflows, not isolated architectures: original resizing, local resolution (384 versus 1008), patch size, descriptor width, and calibration coverage differ. SAM3 calibration includes only original/10\%/5\%, whereas DINOv3 uses broader distinct crop sizes.

The SAM3 gain comes mainly from recall, which rises from 0.1559 to 0.8012 and 0.9230, while precision changes from 0.3403 to 0.3083 and 0.3227. Boundary density rises from 0.0445 to 0.4064 and 0.5895. Tolerant matching permits multiple predicted pixels per reference edge, so these dense partitions do not establish cleaner objects or calibrated boundary certainty. In the three-axis sensitivity, SAM3 means are 0.2654, 0.5317, and 0.5284: 5\% does \emph{not} improve over 10\%. The scale response depends on retained representation.

The fungal illustration exposes narrow branching outlines blurred in the original SAM3 display. Its primary F1 is 0.1258, 0.3040, and 0.3439; recall is 0.0982, 0.6908, and 0.9109. These are cues consistent with the photographed network, not annotated individual hyphae. Substrate, illumination, and the central inoculum also contribute. No hypha-detection or biological segmentation accuracy is claimed.

\FloatBarrier

\FloatBarrier

\begingroup\small
\bibliographystyle{unsrtnat} \bibliography{references}
\endgroup
\end{document}